\documentclass[manuscript,screen]{acmart} 

\usepackage{multirow}
\usepackage{amsmath}
\usepackage{relsize}
\usepackage{graphicx}
\usepackage{listings}
\usepackage{xcolor}
\usepackage{booktabs}
\usepackage{fancyref}
\usepackage{import}
\usepackage{dsfont}
\usepackage{lineno}
\usepackage{float}
\usepackage{caption}
\usepackage[skip=10pt]{subcaption}
\usepackage{bm}
\usepackage{bbm}
\usepackage{commath}
\usepackage{tikz}

\usepackage{fancyhdr}
\usepackage{datetime}

\hypersetup{pdfinfo={
  CreationDate={D:20200407195600},
  ModDate={D:20200407195600}
}}

\usepackage{algorithm}
\usepackage{algorithmic}
\usepackage{enumitem}
\usetikzlibrary{calc}
\usetikzlibrary{bayesnet}
\usepackage{pythontex} 

\usepackage{wrapfig} 
\usepackage{color} 
\definecolor{darkgreen}{RGB}{0,179,0}

\newcommand{\myref}[2]{\hyperref[#2]{#1 \ref*{#2}}}

\newboolean{neurips}
\setboolean{neurips}{false}

\usepackage{amsthm}

\theoremstyle{definition} 

\hypersetup{pdfinfo={
    CreationDate={D:20200407195600},
    ModDate={D:20200407195600}
}}

\newdate{date}{03}{10}{2021}

\begin{document}

\title{Discovering Supply Chain Links with Augmented Intelligence}

\author{Achintya Gopal}
\email{agopal6@bloomberg.net}
\affiliation{%
  \institution{Bloomberg}
  \streetaddress{731 Lexington Avenue}
  \city{New York}
  \state{Ny}
  \country{USA}
  \postcode{10022}
}

\author{Chunho Chang$^*$}
\email{chunhochang1015@gmail.com}
\affiliation{%
  \institution{Two Sigma}
  \streetaddress{100 6th Ave}
  \city{New York}
  \state{Ny}
  \country{USA}
  \postcode{10013}
}

\newcommand\blfootnote[1]{%
  \begingroup
  \renewcommand\thefootnote{}\footnotetext{#1}%
  \addtocounter{footnote}{0}%
  \endgroup
}
\blfootnote{$^*$ Work done while at Bloomberg}

\date{\displaydate{date}}

\begin{abstract}
One of the key components in analyzing the risk of a company is understanding a company's supply chain. Supply chains are constantly disrupted, whether by tariffs, pandemics, severe weather, etc. In this paper, we tackle the problem of predicting previously unknown suppliers and customers of companies using graph neural networks (GNNs) and show strong performance in finding previously unknown connections by combining the predictions of our model and the domain expertise of supply chain analysts.
\end{abstract}

\maketitle

\section{Introduction}

A key component in analyzing the risk of a company is understanding a company's supply chain. Supply chains are constantly disrupted, whether by tariffs \citep{Grossman2021TariffsSPLC}, pandemics \citep{Aca2020PandemicSPLC}, climate change \citep{Ghadge2019ClimateChangeSPLC}, etc. Understanding the risk attributed to such supply chain disruptions is crucial in ensuring safe investments. Though there are regulations in some countries requiring reporting of supply chain relationships \citep{FASBSPLC}, these regulations do not require \textit{all} to be reported and many go unannounced. Hence, we need methods to help shed light on likely connections; in this paper, we use machine learing to help with this task.

Though often the goal of artificial intelligence (AI) is posed as replacing humans, in the majority of problems, this goal is still far out of reach. Due to this, a more realistic goal is \textit{augmented intelligence}, ``increasing the capability of a man to approach a complex problem situation, to gain comprehension to suit his particular needs, and to derive solutions to problems.'' \citep{Englebart1962AugmentingHI}. In other words, the goal is a partnership between people and artificial intelligence, working together to combine the strengths of both in order to improve decision making. 

Though machines are well-suited to processing large amounts of data and learning statistical relationships from data, there can be (and more often are) nuances and causal relationships that a human is much better at sifting through. It is these two strengths that we aim to combine in this paper to find previously unknown customers.\footnote{In this paper, we focus our analysis on predicting customers though the model can trivially be extended to suppliers.} We achieve this goal by:

\begin{enumerate}

\item First, training a Graph Neural Network (GNN) (\myref{Section}{sec:model}) on our Supply Chain dataset (\myref{Section}{sec:splc_data}) as well as many other company-level datasets (\myref{Section}{sec:data}) to learn to recognize situations where there is a high chance that a company is trading with another, though we have not found evidence yet.

\item Second, taking the predictions from our GNN, we search for evidence in primary sources for the existence of these new edges (\myref{Section}{sec:evaluation_human}).

\end{enumerate}

\section{Terminology}

\begin{itemize}
    \item \textit{Positive/negative edge} An edge in the supply chain dataset refers to a supplier-customer connection between two companies. We refer to this as a positive edge since, in the binary classification context, we can view the existence of an edge as the positive class; a negative edge in the supply chain dataset refers to a supplier-customer connection that \textit{does not exist} between two companies. 

    \item \textit{Number of hops} We use the word \textit{hops} to denote the distance between two nodes where distance is defined by the minimum number of edges needed to connect the nodes. For example, nodes that are one-hop away are the neighbors of a node, i.e. the suppliers and customers of a company; nodes that are two-hops away are the neighbors of the neighbors, i.e. suppliers and customers of the suppliers and customers of a company.

\end{itemize}

\section{Supply Chain Data}\label{sec:splc_data}

Our Supply Chain dataset covers a global set of over 23,000 companies and provides information on the supplier-customer relationships between those companies and others. Since each company can have multiple relationships, there are over 250,000 companies in the dataset in total. These connections were found from primary sources such as company presentations.

\begin{figure}[!t]
  \begin{minipage}[b]{0.48\textwidth}
    \centering
    \centerline{\includegraphics[width=\columnwidth]{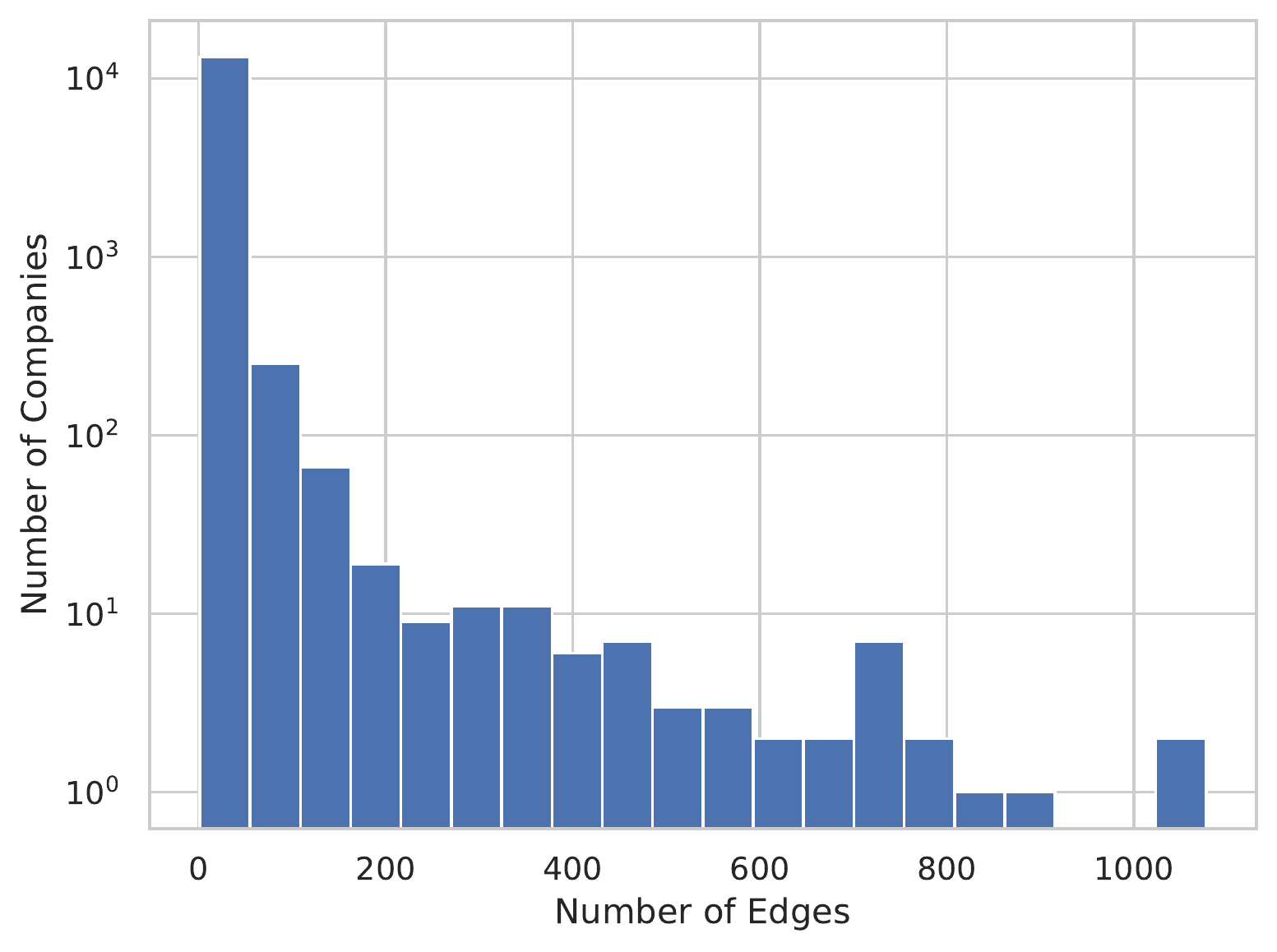}}
    \captionof{figure}{Number of edges per US company for 2020 supply chain data.}\label{fig:usd_coverage}
  \end{minipage}
  \hfill
  \begin{minipage}[b]{0.48\textwidth}
    \centering
      \begin{tabular}{lccc}
  \toprule
   & \multicolumn{3}{c}{Coverage on Customers}
  \\
   \cmidrule(lr){2-4}
  Country &  Average  &  Median  &  Max  \\
  \midrule
  China              &                  5.33 &                       1 &                  617 \\
  United States      &                  7.98 &                       1 &                 1077 \\
  Japan              &                  7.60 &                       1 &                 1137 \\
  Korea              &                  7.65 &                       2 &                  358 \\
  India              &                  5.44 &                       1 &                  285 \\
  \bottomrule
  \end{tabular}
  \captionof{table}{Supply chain coverage statistics for 2020. The five countries with the highest number of companies are shown. We also present the average/median/max number of edges (customers) per company in the country.}\label{tbl:splc_edge_counts}
    \end{minipage}
  \end{figure}

In \myref{Table}{tbl:splc_edge_counts}, we can see that the average number of edges per company per country is much larger than the median number of edges; this implies a highly skewed distribution where the majority of companies have very few observed edges. For example, 63\% of US companies only have one reported edge; however, there is one US company where 1077 customers are covered. This is seen in \myref{Figure}{fig:usd_coverage} where the y-axis is log scale, showing that majority of companies have very few reported customers. The impact of the skew in terms of evaluation will be revisted in the construction of the validation set (\myref{Section}{sec:eval_validation}).

Importantly, the fact that we have very few edges for the majority of companies motivates the importance of the problem; without broader insight into the supply chains of these companies, the risk will be unclear.

\section{Company-Level Data}\label{sec:data}

Though many methods in the literature approach link prediction using a transductive method (e.g. \cite{bordes2013TransE,Kipf2017GCN}), we use an inductive approach. In other words, we use features (external information) for each company instead of using only an embedding given the company ID. One reason for this is that it allows our model to generalize to companies it has never seen before. Further, the intuition behind adding features to the model is that it gives the model hints on notions of similarity.

In order to help our model be able to make strong specific predictions, in addition to the Supply Chain dataset (\myref{Section}{sec:splc_data}), we included datasets domain experts currently use in their analysis of edge predictions as well as datasets they believe would be useful. An important method in deciding datasets is to not only include the information that domain experts currently use, but to include finer grained information of that data that humans might have a harder time learning relationships from. For example, one methodology used by domain experts is to analyze the industry, location, and size of companies. When this information is insufficient or missing, domain experts might further use company description and competitors.

Using this, we included industry classification, revenue segmentation (percentage of revenue from each industry sector), locations (where the company is located as well as location of facilities), company financials (balance sheet, income statement, and statement of cash flows), raw materials the company makes, company description (textual data), and competitors.


\section{Model Methodology}\label{sec:model}

In this section, we break down our model methodology into two parts: the model architecture (\myref{Section}{sec:model_arch}) and the model training (\myref{Section}{sec:model_training}). At a high level, we use graph attention networks as our model architecture and used a two-step training procedure where the first step focuses on recall and the second focuses on precision.

\subsection{Model Architecture}\label{sec:model_arch}

A common theme in the success of deep learning has been in the finding and usage of models that are well-catered to the structure of the data, e.g. recurrent neural networks for time series \citep{hochreiter1997long,Cho2014GRU,DeepAR2017}, attention for language \citep{vaswani2017attention,devlin2018bert,radford2018improving}, and convolutional networks for images \citep{he2016resnet,Alhashim2018DenseDepth}. Supply chain and competitors data is easiest to represent in the form of a graph.
Among the methods that exist to process graphs, Graph Attention Networks \citep{Velickovic2018model} best fit our needs.

Our model can be decomposed into four components: 
\begin{enumerate}
    \item \textit{Feature MLP}: We first apply a multilayer perceptron (MLP) on the features of each node in the graph, aggregating the many different datasets we are using (\myref{Section}{sec:data}). Though most of the data we utilized is numerical, company description is not; we processed (embedded) company description using BERT \citep{devlin2018bert}. A more thorough analysis of BERT and its variants can be found in \myref{Appendix}{sec:bert}.

    \item \textit{Neighbor aggregation}: The specific formulation of the graph attentional layer (GAT) can be found in \myref{Appendix}{sec:gat_layer}. 
    At a high level, the layer takes a graph and representations $\vec{h}_i$ for each node $i$ as input and outputs a new representation $\vec{h'}_{i, e}$ ($e$ denoting the edge type) for each node where the new representation is formed by using an attention-based function whose inputs are the previous representation of the node and the previous representations of the neighbors of the node. 
    To handle multiple edge types, we apply the graph attention layer (\myref{Equation}{eqn:gat_layer}) per edge type (i.e. once for suppliers $\vec{h'}_{i,s}$, once for customers $\vec{h'}_{i,c}$, and once for competitors $\vec{h'}_{i,t}$). To aggregate this information, we concatenate the outputs for each edge type along with the node features and apply an MLP to aggregate the information:
   $$ \vec{h'}_i = f_\theta(\vec{h'}_{i,s}\ ||\ \vec{h'}_{i,c}\ ||\ \vec{h'}_{i,t}\ ||\ \vec{h}_{i})$$
where $||$ denotes concatenation and $f_\theta$ is an MLP.
    \item \textit{Aggregating multiple hops}: To handle multiple hops, we apply a GAT layer for each hop away. For example, for a two-hop model, we first apply a GAT layer for nodes two-hops away; using the outputs from that layer, we apply another GAT layer one-hop away; finally, we aggregate that information using a GAT layer at the node in question.

    \item \textit{Scoring a pair of companies}: Using the architecture above, we apply it to the supplier side (company A) and customer side (company B) to get representations $e_{a}$ and $e_{b}$, respectively. Given the two embeddings, we generate a score for how likely there is a supplier-customer edge between the two by taking the dot product of the two: $s_{a,b} = e_a \cdot e_b$. 
    By using a different network for suppliers and a different network for customers, we allow for asymmetry in prediction (company B being company A's supplier does not imply company A is company B's supplier).

\end{enumerate}

\subsection{Model Training}\label{sec:model_training}

The main difficulties in link prediction comes down to two things: there are no negative edges (we do not know with certainty which edges do not exist) and highly imbalanced classes (there is a significantly larger number of negative edges than positive edges).

Though there is no rigorous solution to the first problem, we simply sample random company pairs from unconnected companies and denote them as a negative edge. We argue this works since most company pairs are not valid relationships (the second issue), meaning if we randomly sample a pair of companies, there is a very low probability that the random pair is a valid supply chain relationship.

Taken from the link prediction literature, we use ``pairwise logistic loss'' \citep{Mohamed2019LossFI}.
Given a positive pair (A, B), we randomly sample a new company C to create a negative pair (A, C). We then score the pairs using the architecture described in the previous section (\myref{Section}{sec:model_arch}) to get scores $s_{a,b}$ and $s_{a,c}$. Given these scores, the loss function is:
\begin{equation}\label{eqn:our_loss}
\log \left( 1 + e^{s_{a,b} - s_{a,c}} \right)
\end{equation}
One interpretation of the sampling method used for this loss function is that, though there are significantly more negative examples than positive ones, we resample such that the two classes are balanced.

Though there is a large body of literature on different loss functions for knowledge graph models \citep{nickel2011RESCAL,Nickel2016HolE,Trouillon2016ComplEX,bordes2013TransE,yang2015DistMult,Lacroix2018Canonical,Dettmers2018ConvE}, we leave it to future work to explore more variants; however, preliminary experiments showed that RankNet \citep{Burges2005RankNET} and LambdaRANK \citep{Burges2006LambdaRANK} hurt performance.

One issue with the training procedure described above is the imbalanced classes.
The main issue with imbalanced classes is that neural networks might struggle to learn both obvious relationships (the relationships that can easily be noticed to be a negative edge) and specific relationships (the nuances that are required to reject certain relationships).

To fix this problem, we use a \textit{two-step model}: first, we train a model where the negative pairs are created by sampling randomly; second, we train another model from scratch where the negative pairs used are pairs that the first model gives a large score to. ``Large score'' here means the score is larger than some predefined threshold. This method can be viewed as boosting where the second model boosts performance by fixing the mistakes of the first model.

\section{Model Validation}

\subsection{Validation Set}\label{sec:eval_validation}

In order to validate our model, we split the observed edges (supplier-customer relationships) into a training and validation set. However, we do not randomly split the edges as this would lead our validation set to contain more edges from companies that already have a lot of edges. As mentioned in \myref{Section}{sec:splc_data}, there are some companies with hundreds of known customers. In terms of evaluation, we care more about our model's performance on companies with only a few observed edges; due to this, we create our validation set by sampling two edges per company for companies that we have no more than twenty known customers. Further, since we trained our model on historical supply chain data, to ensure no data leakage, for every edge in the validation set, we removed it from the training set across all years .

\subsection{Evaluation Metrics}

To evaluate our model, for each company in our validation set, we score every other company with respect to it. Using these scores, we get the rank for each company; importantly, we filter out pairs that were in the training set before computing the ranks since the ranks for training edges are generally higher than for other edges. 

To evaluate, we used the mean rank, recall@N (percentage of edges in the validation set whose rank is less than N), and hit@N (percentage of companies that have \textit{at least one} positive edge in the top N).
Note in the last two metrics, the percentages get larger as N grows. Further, since there are many edges we do not know of, the recall@N and hit@N are lower bounds on the true value. For example, if the hit@20 is 40\% (meaning that 40\% of the companies have at least one true edge in the top 20), for the other 60\% companies, there might be true edges in the top 20 that we currently do not know of.

Though there can sometimes be tradeoffs with respect to the three metrics, the metric we gave priority to was hit@20 because the goal (\myref{Section}{sec:evaluation_human}) is to find previously unknown edges.

\subsection{Evaluation}\label{sec:evaluation}
\begin{table}[!bt]
\begin{center}
\begin{tabular}{lcccc}
\toprule
  & Mean Rank ($\downarrow$) &  Recall@100 ($\uparrow$) & Hit@20 ($\uparrow$) \\
\midrule
Baseline & 2246 & 25\%& 16\% \\
+ Facilities & 2058  & 27\%& 18\% \\
+ Fundamentals + Raw & 1854 & 32\%& 22\% \\
+ BERT & 1832  & 31\%& 22\% \\
- BERT + Graph & 947  & 53\% & 40\% \\
\midrule
+ Two step (Final) & \textbf{938} & \textbf{56\%} & \textbf{45\%} \\
\bottomrule
\end{tabular}
\end{center}
\caption{Evaluation of our model given different amounts of data. The baseline model contains industry classification, segmentation, and country data. The models with graph information contains all the features before it (except BERT).}
\label{fig:evaluation}
\end{table}


In \myref{Figure}{fig:evaluation}, we show the performance on our validation set given different amounts of data. Our baseline model contains industry classification, segmentation, and country information. As can be seen, facilities gives a reasonable boost over the information contained in country data; further, adding fundamentals and raw materials (more information about the business of the company) also gives a reasonable boost in performance. However, BERT hurts performance suggesting that the information in the company description is encompassed by the features already contained in the model and that adding BERT has simply lead to overfitting; though BERT hurts performance, \myref{Appendix}{sec:bert} shows that, for companies with limited data, BERT (company description) can lend valuable information. The biggest boost in performance comes from adding supply chain and competitors data where, to incorporate this information, we tuned the number of hops. Further, training a second step model, though not a big improvement, gives a measurable performance boost.

\section{Finding Previously Unknown Links}\label{sec:evaluation_human}

As noted previously, since there are many edges that we do not know of, the hit@20 is a lower bound on the true value. To get a better understanding of the true hit@20, we gave the top twenty predictions of all companies from our final model to supply chain analysts (domain experts) to see if they can find evidence for the predictions. 
For our analysis, supply chain experts selected ten industries where the current coverage has enough observations for the model to have learned something meaningful while not so many where most of the publicly available relations are previously found.
From these industries, twenty predictions were sampled randomly and assigned to experts in those industries. In total, 180 predictions were analyzed by nine experts. In addition to these predictions, we gave two Google search URLs: customer name from supplier's website and vice versa.

\begin{table}[!bt]
\begin{center}
\begin{tabular}{lcccc}
\toprule
Country & Total & Found  & \% Found  & Found by Google \\
\midrule
United States & 33 & 13& 39\% & 9 \\
China & 29 & 4& 14\% & 0 \\
Japan & 17 & 2& 12\% & 1 \\
India & 13 & 2& 15\% & 2 \\
Taiwan & 11 & 1 & 9\% & 1 \\
\midrule
Total & 180 & 37& 21\% & 22 \\
\bottomrule
\end{tabular}
\end{center}
\caption{Results of analysis by domain experts. The results shown are grouped by supplier-side country and the five countries with the most edges analyzed are shown.}\label{tbl:human_eval}
\end{table}

In \myref{Table}{tbl:human_eval}, we see that evidence was found for approximately 21\% of the edges evaluated; in other words, the hit@10 is approximately 91\%\footnote{$0.91 = 1 - (1 - 0.21)^{10}$, assuming the ability to find an edge is independent of the supplier and customer}. Thus, whereas our hit@10 for our model is 36\%, directly measuring this statistic using predictions and external sources shows our model is much better than the statistics show (as is expected from the fact that we knew our metrics would be lower bounds of the real value).

Further, if we focus on results where the supplier is a US company, nearly 39\% of the relationships were found. This percentage is expected to be higher as the ability to find evidence for edges can sometimes be predicated upon knowing the language of the country in which the companies are in.
In terms of the usefulness of the Google links given, 60\% of the edges found were found from one of the two Google links given and nearly 70\% of the US edges found were found from one of the Google links. 

In conclusion, combining these observations together, a simple method to find new edges with primary evidence is taking the top twenty predictions of our model for US companies and searching only two Google links. Further, these results also validate the overall quality of our model, and that, even if primary evidence cannot be found, the predictions can be trusted.

\newpage
\bibliographystyle{chicago}
\bibliography{./biblio}

\begin{thebibliography}{}

\bibitem[\protect\citeauthoryear{{Alhashim} and {Wonka}}{{Alhashim} and
  {Wonka}}{2018}]{Alhashim2018DenseDepth}
{Alhashim}, I. and P.~{Wonka} (2018, December).
\newblock {High Quality Monocular Depth Estimation via Transfer Learning}.
\newblock {\em arXiv e-prints\/}, arXiv:1812.11941.

\bibitem[\protect\citeauthoryear{An, Chen, Han, and Sun}{An
  et~al.}{2018}]{an2018accurate}
An, B., B.~Chen, X.~Han, and L.~Sun (2018).
\newblock Accurate text-enhanced knowledge graph representation learning.
\newblock In {\em NAACL}, pp.\  745--755.

\bibitem[\protect\citeauthoryear{Ando and Zhang}{Ando and
  Zhang}{2005}]{ando2005words}
Ando, R.~K. and T.~Zhang (2005, December).
\newblock A framework for learning predictive structures from multiple tasks
  and unlabeled data.
\newblock {\em J. Mach. Learn. Res.\/}~{\em 6}, 1817–1853.

\bibitem[\protect\citeauthoryear{Atwood and Towsley}{Atwood and
  Towsley}{2016}]{Atwood2016Diffusion}
Atwood, J. and D.~Towsley (2016).
\newblock Diffusion-convolutional neural networks.
\newblock In D.~Lee, M.~Sugiyama, U.~Luxburg, I.~Guyon, and R.~Garnett (Eds.),
  {\em Advances in Neural Information Processing Systems}, Volume~29. Curran
  Associates, Inc.

\bibitem[\protect\citeauthoryear{Bahdanau, Cho, and Bengio}{Bahdanau
  et~al.}{2015}]{bahdanau2014neural}
Bahdanau, D., K.~Cho, and Y.~Bengio (2015, January).
\newblock Neural machine translation by jointly learning to align and
  translate.
\newblock 3rd International Conference on Learning Representations, ICLR 2015 ;
  Conference date: 07-05-2015 Through 09-05-2015.

\bibitem[\protect\citeauthoryear{Blitzer, McDonald, and Pereira}{Blitzer
  et~al.}{2006}]{blitzer2006domainadap}
Blitzer, J., R.~McDonald, and F.~Pereira (2006, July).
\newblock Domain adaptation with structural correspondence learning.
\newblock In {\em Proceedings of the 2006 Conference on Empirical Methods in
  Natural Language Processing}, Sydney, Australia, pp.\  120--128. Association
  for Computational Linguistics.

\bibitem[\protect\citeauthoryear{Bordes, Usunier, Garcia-Duran, Weston, and
  Yakhnenko}{Bordes et~al.}{2013}]{bordes2013TransE}
Bordes, A., N.~Usunier, A.~Garcia-Duran, J.~Weston, and O.~Yakhnenko (2013).
\newblock Translating embeddings for modeling multi-relational data.
\newblock In {\em NIPS}, pp.\  2787--2795.

\bibitem[\protect\citeauthoryear{Brown, Della~Pietra, deSouza, Lai, and
  Mercer}{Brown et~al.}{1992}]{brown1992ngram}
Brown, P.~F., V.~J. Della~Pietra, P.~V. deSouza, J.~C. Lai, and R.~L. Mercer
  (1992).
\newblock Class-based \textit{n}-gram models of natural language.
\newblock {\em Computational Linguistics\/}~{\em 18\/}(4), 467--480.

\bibitem[\protect\citeauthoryear{Brown, Mann, Ryder, Subbiah, Kaplan, Dhariwal,
  Neelakantan, Shyam, Sastry, Askell, et~al.}{Brown
  et~al.}{2020}]{brown2020language}
Brown, T.~B., B.~Mann, N.~Ryder, M.~Subbiah, J.~Kaplan, P.~Dhariwal,
  A.~Neelakantan, P.~Shyam, G.~Sastry, A.~Askell, et~al. (2020).
\newblock Language models are few-shot learners.
\newblock {\em arXiv preprint arXiv:2005.14165\/}.

\bibitem[\protect\citeauthoryear{Bruna, Zaremba, Szlam, and LeCun}{Bruna
  et~al.}{2014}]{Bruna2013Spectral}
Bruna, J., W.~Zaremba, A.~Szlam, and Y.~LeCun (2014).
\newblock Spectral networks and locally connected networks on graphs.
\newblock In Y.~Bengio and Y.~LeCun (Eds.), {\em 2nd International Conference
  on Learning Representations, {ICLR} 2014, Banff, AB, Canada, April 14-16,
  2014, Conference Track Proceedings}.

\bibitem[\protect\citeauthoryear{Burges, Ragno, and Le}{Burges
  et~al.}{2007}]{Burges2006LambdaRANK}
Burges, C., R.~Ragno, and Q.~Le (2007).
\newblock Learning to rank with nonsmooth cost functions.
\newblock In B.~Sch\"{o}lkopf, J.~Platt, and T.~Hoffman (Eds.), {\em Advances
  in Neural Information Processing Systems}, Volume~19. MIT Press.

\bibitem[\protect\citeauthoryear{Burges, Shaked, Renshaw, Lazier, Deeds,
  Hamilton, and Hullender}{Burges et~al.}{2005}]{Burges2005RankNET}
Burges, C., T.~Shaked, E.~Renshaw, A.~Lazier, M.~Deeds, N.~Hamilton, and
  G.~Hullender (2005).
\newblock Learning to rank using gradient descent.
\newblock In {\em Proceedings of the 22nd International Conference on Machine
  Learning}, ICML '05, New York, NY, USA, pp.\  89–96. Association for
  Computing Machinery.

\bibitem[\protect\citeauthoryear{Cho, {van Merrienboer}, Gulcehre, Bougares,
  Schwenk, and Bengio}{Cho et~al.}{2014}]{Cho2014GRU}
Cho, K., B.~{van Merrienboer}, C.~Gulcehre, F.~Bougares, H.~Schwenk, and
  Y.~Bengio (2014).
\newblock Learning phrase representations using rnn encoder-decoder for
  statistical machine translation.
\newblock In {\em Conference on Empirical Methods in Natural Language
  Processing (EMNLP 2014)}.

\bibitem[\protect\citeauthoryear{Choi, Kim, Joe, and Gwon}{Choi
  et~al.}{2021}]{Choi2020EvalBERT}
Choi, H., J.~Kim, S.~Joe, and Y.~Gwon (2021).
\newblock Evaluation of bert and albert sentence embedding performance on
  downstream nlp tasks.
\newblock In {\em 2020 25th International Conference on Pattern Recognition
  (ICPR)}, pp.\  5482--5487.

\bibitem[\protect\citeauthoryear{Defferrard, Bresson, and
  Vandergheynst}{Defferrard et~al.}{2016}]{Defferrard2016Convolutional}
Defferrard, M., X.~Bresson, and P.~Vandergheynst (2016).
\newblock Convolutional neural networks on graphs with fast localized spectral
  filtering.
\newblock In D.~Lee, M.~Sugiyama, U.~Luxburg, I.~Guyon, and R.~Garnett (Eds.),
  {\em Advances in Neural Information Processing Systems}, Volume~29. Curran
  Associates, Inc.

\bibitem[\protect\citeauthoryear{Dettmers, Minervini, Stenetorp, and
  Riedel}{Dettmers et~al.}{2018}]{Dettmers2018ConvE}
Dettmers, T., P.~Minervini, P.~Stenetorp, and S.~Riedel (2018, Apr.).
\newblock Convolutional 2d knowledge graph embeddings.
\newblock {\em Proceedings of the AAAI Conference on Artificial
  Intelligence\/}~{\em 32\/}(1).

\bibitem[\protect\citeauthoryear{Devlin, Chang, Lee, and Toutanova}{Devlin
  et~al.}{2019}]{devlin2018bert}
Devlin, J., M.-W. Chang, K.~Lee, and K.~Toutanova (2019, June).
\newblock {BERT}: Pre-training of deep bidirectional transformers for language
  understanding.
\newblock In {\em Proceedings of the 2019 Conference of the North {A}merican
  Chapter of the Association for Computational Linguistics: Human Language
  Technologies, Volume 1 (Long and Short Papers)}, Minneapolis, Minnesota, pp.\
   4171--4186. Association for Computational Linguistics.

\bibitem[\protect\citeauthoryear{Duvenaud, Maclaurin, Iparraguirre, Bombarell,
  Hirzel, Aspuru-Guzik, and Adams}{Duvenaud
  et~al.}{2015}]{Duvenaud2015Convolution}
Duvenaud, D.~K., D.~Maclaurin, J.~Iparraguirre, R.~Bombarell, T.~Hirzel,
  A.~Aspuru-Guzik, and R.~P. Adams (2015).
\newblock Convolutional networks on graphs for learning molecular fingerprints.
\newblock In C.~Cortes, N.~Lawrence, D.~Lee, M.~Sugiyama, and R.~Garnett
  (Eds.), {\em Advances in Neural Information Processing Systems}, Volume~28.
  Curran Associates, Inc.

\bibitem[\protect\citeauthoryear{Englebart}{Englebart}{1962}]{Englebart1962AugmentingHI}
Englebart, D. (1962).
\newblock Augmenting human intellect: a conceptual framework.

\bibitem[\protect\citeauthoryear{FASB}{FASB}{}]{FASBSPLC}
FASB.
\newblock Fasb statement no. 14 financial reporting for segments of a business
  enterprise.

\bibitem[\protect\citeauthoryear{Feng, Huang, Yang, and Zhu}{Feng
  et~al.}{2016}]{Feng2016GAKE}
Feng, J., M.~Huang, Y.~Yang, and X.~Zhu (2016, December).
\newblock {GAKE}: Graph aware knowledge embedding.
\newblock In {\em Proceedings of {COLING} 2016, the 26th International
  Conference on Computational Linguistics: Technical Papers}, Osaka, Japan,
  pp.\  641--651. The COLING 2016 Organizing Committee.

\bibitem[\protect\citeauthoryear{Frasconi, Gori, and Sperduti}{Frasconi
  et~al.}{1998}]{Frasconi1998GNN}
Frasconi, P., M.~Gori, and A.~Sperduti (1998).
\newblock A general framework for adaptive processing of data structures.
\newblock {\em IEEE Transactions on Neural Networks\/}~{\em 9\/}(5), 768--786.

\bibitem[\protect\citeauthoryear{Ghadge, Hendrik, and Seuring}{Ghadge
  et~al.}{2019}]{Ghadge2019ClimateChangeSPLC}
Ghadge, D.~A., W.~Hendrik, and S.~Seuring (2019, 06).
\newblock Managing climate change risks in global supply chains: A review and
  research agenda.
\newblock {\em International Journal of Production Research\/}~{\em 58}.

\bibitem[\protect\citeauthoryear{Gori, Monfardini, and Scarselli}{Gori
  et~al.}{2005}]{Gori2005GNN}
Gori, M., G.~Monfardini, and F.~Scarselli (2005).
\newblock A new model for learning in graph domains.
\newblock In {\em Proceedings. 2005 IEEE International Joint Conference on
  Neural Networks, 2005.}, Volume~2, pp.\  729--734 vol. 2.

\bibitem[\protect\citeauthoryear{Grossman and Helpman}{Grossman and
  Helpman}{2021}]{Grossman2021TariffsSPLC}
Grossman, G.~M. and E.~Helpman (2021, January).
\newblock {When Tariffs Disrupt Global Supply Chains}.
\newblock Working Papers 274, Princeton University, Department of Economics,
  Center for Economic Policy Studies.

\bibitem[\protect\citeauthoryear{Grover and Leskovec}{Grover and
  Leskovec}{2016}]{grover2016node2vec}
Grover, A. and J.~Leskovec (2016).
\newblock Node2vec: Scalable feature learning for networks.
\newblock In {\em Proceedings of the 22nd ACM SIGKDD International Conference
  on Knowledge Discovery and Data Mining}, KDD '16, New York, NY, USA, pp.\
  855–864. Association for Computing Machinery.

\bibitem[\protect\citeauthoryear{Guu, Miller, and Liang}{Guu
  et~al.}{2015}]{Guu2015Traversing}
Guu, K., J.~Miller, and P.~Liang (2015, September).
\newblock Traversing knowledge graphs in vector space.
\newblock In {\em Proceedings of the 2015 Conference on Empirical Methods in
  Natural Language Processing}, Lisbon, Portugal, pp.\  318--327. Association
  for Computational Linguistics.

\bibitem[\protect\citeauthoryear{Hamilton, Ying, and Leskovec}{Hamilton
  et~al.}{2017a}]{Hamilton2017GraphSAGE}
Hamilton, W., Z.~Ying, and J.~Leskovec (2017a).
\newblock Inductive representation learning on large graphs.
\newblock In I.~Guyon, U.~V. Luxburg, S.~Bengio, H.~Wallach, R.~Fergus,
  S.~Vishwanathan, and R.~Garnett (Eds.), {\em Advances in Neural Information
  Processing Systems}, Volume~30. Curran Associates, Inc.

\bibitem[\protect\citeauthoryear{Hamilton, Ying, and Leskovec}{Hamilton
  et~al.}{2017b}]{Hamilton2017model}
Hamilton, W.~L., R.~Ying, and J.~Leskovec (2017b).
\newblock Inductive representation learning on large graphs.
\newblock pp.\  1024--1034. NIPS 2017.

\bibitem[\protect\citeauthoryear{He, Zhang, Ren, and Sun}{He
  et~al.}{2016}]{he2016resnet}
He, K., X.~Zhang, S.~Ren, and J.~Sun (2016).
\newblock Deep residual learning for image recognition.
\newblock In {\em Proceedings of the IEEE conference on computer vision and
  pattern recognition}, pp.\  770--778.

\bibitem[\protect\citeauthoryear{Henaff, Bruna, and LeCun}{Henaff
  et~al.}{2015}]{Henaff2015DeepCN}
Henaff, M., J.~Bruna, and Y.~LeCun (2015).
\newblock Deep convolutional networks on graph-structured data.
\newblock {\em ArXiv\/}~{\em abs/1506.05163}.

\bibitem[\protect\citeauthoryear{Hochreiter and Schmidhuber}{Hochreiter and
  Schmidhuber}{1997}]{hochreiter1997long}
Hochreiter, S. and J.~Schmidhuber (1997).
\newblock Long short-term memory.
\newblock {\em Neural computation\/}~{\em 9\/}(8), 1735--1780.

\bibitem[\protect\citeauthoryear{Jawahar, Sagot, and Seddah}{Jawahar
  et~al.}{2019}]{jawahar2019bert}
Jawahar, G., B.~Sagot, and D.~Seddah (2019, July).
\newblock What does {BERT} learn about the structure of language?
\newblock In {\em Proceedings of the 57th Annual Meeting of the Association for
  Computational Linguistics}, Florence, Italy, pp.\  3651--3657. Association
  for Computational Linguistics.

\bibitem[\protect\citeauthoryear{Jiang, Tresp, Huang, and Nickel}{Jiang
  et~al.}{2012}]{Jiang2021MultiRelGraph}
Jiang, X., V.~Tresp, Y.~Huang, and M.~Nickel (2012).
\newblock Link prediction in multi-relational graphs using additive models.
\newblock In {\em Proceedings of the 2012 International Conference on Semantic
  Technologies Meet Recommender Systems \& Big Data - Volume 919}, SeRSy'12,
  Aachen, DEU, pp.\  1–12. CEUR-WS.org.

\bibitem[\protect\citeauthoryear{Kipf and Welling}{Kipf and
  Welling}{2017}]{Kipf2017GCN}
Kipf, T.~N. and M.~Welling (2017).
\newblock Semi-supervised classification with graph convolutional networks.
\newblock In {\em 5th International Conference on Learning Representations,
  {ICLR} 2017, Toulon, France, April 24-26, 2017, Conference Track
  Proceedings}. OpenReview.net.

\bibitem[\protect\citeauthoryear{Kiros, Zhu, Salakhutdinov, Zemel, Urtasun,
  Torralba, and Fidler}{Kiros et~al.}{2015}]{Kiros2015Skip}
Kiros, R., Y.~Zhu, R.~R. Salakhutdinov, R.~Zemel, R.~Urtasun, A.~Torralba, and
  S.~Fidler (2015).
\newblock Skip-thought vectors.
\newblock In C.~Cortes, N.~Lawrence, D.~Lee, M.~Sugiyama, and R.~Garnett
  (Eds.), {\em Advances in Neural Information Processing Systems}, Volume~28.
  Curran Associates, Inc.

\bibitem[\protect\citeauthoryear{Lacroix, Usunier, and Obozinski}{Lacroix
  et~al.}{2018}]{Lacroix2018Canonical}
Lacroix, T., N.~Usunier, and G.~Obozinski (2018, 10--15 Jul).
\newblock Canonical tensor decomposition for knowledge base completion.
\newblock In J.~Dy and A.~Krause (Eds.), {\em Proceedings of the 35th
  International Conference on Machine Learning}, Volume~80 of {\em Proceedings
  of Machine Learning Research}, pp.\  2863--2872. PMLR.

\bibitem[\protect\citeauthoryear{Lan, Chen, Goodman, Gimpel, Sharma, and
  Soricut}{Lan et~al.}{2019}]{Lan2019ALBERT}
Lan, Z., M.~Chen, S.~Goodman, K.~Gimpel, P.~Sharma, and R.~Soricut (2019).
\newblock {ALBERT:} {A} lite {BERT} for self-supervised learning of language
  representations.
\newblock {\em CoRR\/}~{\em abs/1909.11942}.

\bibitem[\protect\citeauthoryear{Le and Mikolov}{Le and
  Mikolov}{2014}]{quoc2014sentence2vec}
Le, Q. and T.~Mikolov (2014, 22--24 Jun).
\newblock Distributed representations of sentences and documents.
\newblock In E.~P. Xing and T.~Jebara (Eds.), {\em Proceedings of the 31st
  International Conference on Machine Learning}, Volume~32 of {\em Proceedings
  of Machine Learning Research}, Bejing, China, pp.\  1188--1196. PMLR.

\bibitem[\protect\citeauthoryear{Li, Wang, Wang, and Leskovec}{Li
  et~al.}{2020}]{Li2020DistanceEncoding}
Li, P., Y.~Wang, H.~Wang, and J.~Leskovec (2020).
\newblock Distance encoding: Design provably more powerful neural networks for
  graph representation learning.
\newblock In H.~Larochelle, M.~Ranzato, R.~Hadsell, M.~Balcan, and H.~Lin
  (Eds.), {\em Advances in Neural Information Processing Systems 33: Annual
  Conference on Neural Information Processing Systems 2020, NeurIPS 2020,
  December 6-12, 2020, virtual}.

\bibitem[\protect\citeauthoryear{Li, Tarlow, Brockschmidt, and Zemel}{Li
  et~al.}{2016}]{Li2016GatedGNN}
Li, Y., D.~Tarlow, M.~Brockschmidt, and R.~S. Zemel (2016).
\newblock Gated graph sequence neural networks.
\newblock In Y.~Bengio and Y.~LeCun (Eds.), {\em 4th International Conference
  on Learning Representations, {ICLR} 2016, San Juan, Puerto Rico, May 2-4,
  2016, Conference Track Proceedings}.

\bibitem[\protect\citeauthoryear{Lin, Liu, Luan, Sun, Rao, and Liu}{Lin
  et~al.}{2015}]{Lin2015PTransE}
Lin, Y., Z.~Liu, H.~Luan, M.~Sun, S.~Rao, and S.~Liu (2015, September).
\newblock Modeling relation paths for representation learning of knowledge
  bases.
\newblock In {\em Proceedings of the 2015 Conference on Empirical Methods in
  Natural Language Processing}, Lisbon, Portugal, pp.\  705--714. Association
  for Computational Linguistics.

\bibitem[\protect\citeauthoryear{Liu, Gardner, Belinkov, Peters, and Smith}{Liu
  et~al.}{2019}]{liu2019bertlayers}
Liu, N.~F., M.~Gardner, Y.~Belinkov, M.~E. Peters, and N.~A. Smith (2019,
  June).
\newblock Linguistic knowledge and transferability of contextual
  representations.
\newblock In {\em Proceedings of the 2019 Conference of the North {A}merican
  Chapter of the Association for Computational Linguistics: Human Language
  Technologies, Volume 1 (Long and Short Papers)}, Minneapolis, Minnesota, pp.\
   1073--1094. Association for Computational Linguistics.

\bibitem[\protect\citeauthoryear{Liu, Ott, Goyal, Du, Joshi, Chen, Levy, Lewis,
  Zettlemoyer, and Stoyanov}{Liu et~al.}{2019}]{Liu2019RoBERTa}
Liu, Y., M.~Ott, N.~Goyal, J.~Du, M.~Joshi, D.~Chen, O.~Levy, M.~Lewis,
  L.~Zettlemoyer, and V.~Stoyanov (2019).
\newblock Roberta: {A} robustly optimized {BERT} pretraining approach.
\newblock {\em CoRR\/}~{\em abs/1907.11692}.

\bibitem[\protect\citeauthoryear{Logeswaran and Lee}{Logeswaran and
  Lee}{2018}]{logeswaran2018an}
Logeswaran, L. and H.~Lee (2018).
\newblock An efficient framework for learning sentence representations.
\newblock In {\em International Conference on Learning Representations}.

\bibitem[\protect\citeauthoryear{Mikolov, Sutskever, Chen, Corrado, and
  Dean}{Mikolov et~al.}{2013}]{word2vec}
Mikolov, T., I.~Sutskever, K.~Chen, G.~S. Corrado, and J.~Dean (2013).
\newblock Distributed representations of words and phrases and their
  compositionality.
\newblock In C.~J.~C. Burges, L.~Bottou, M.~Welling, Z.~Ghahramani, and K.~Q.
  Weinberger (Eds.), {\em Advances in Neural Information Processing Systems
  26}, pp.\  3111--3119. Curran Associates, Inc.

\bibitem[\protect\citeauthoryear{Mohamed, Nov{\'a}cek, Vandenbussche, and
  Mu{\~n}oz}{Mohamed et~al.}{2019}]{Mohamed2019LossFI}
Mohamed, S.~K., V.~Nov{\'a}cek, P.~Vandenbussche, and E.~Mu{\~n}oz (2019).
\newblock Loss functions in knowledge graph embedding models.
\newblock In {\em DL4KGESWC}.

\bibitem[\protect\citeauthoryear{Monti, Boscaini, Masci, Rodola, Svoboda, and
  Bronstein}{Monti et~al.}{2017}]{Monti2017Geometric}
Monti, F., D.~Boscaini, J.~Masci, E.~Rodola, J.~Svoboda, and M.~M. Bronstein
  (2017, jul).
\newblock Geometric deep learning on graphs and manifolds using mixture model
  cnns.
\newblock In {\em 2017 IEEE Conference on Computer Vision and Pattern
  Recognition (CVPR)}, Los Alamitos, CA, USA, pp.\  5425--5434. IEEE Computer
  Society.

\bibitem[\protect\citeauthoryear{Nickel, Rosasco, and Poggio}{Nickel
  et~al.}{2016}]{Nickel2016HolE}
Nickel, M., L.~Rosasco, and T.~Poggio (2016).
\newblock Holographic embeddings of knowledge graphs.
\newblock In {\em Proceedings of the Thirtieth AAAI Conference on Artificial
  Intelligence}, AAAI'16, pp.\  1955–1961. AAAI Press.

\bibitem[\protect\citeauthoryear{Nickel, Tresp, and Kriegel}{Nickel
  et~al.}{2011}]{nickel2011RESCAL}
Nickel, M., V.~Tresp, and H.-P. Kriegel (2011).
\newblock A three-way model for collective learning on multi-relational data.
\newblock In {\em ICML}, pp.\  809--816.

\bibitem[\protect\citeauthoryear{Pennington, Socher, and Manning}{Pennington
  et~al.}{2014}]{pennington2014glove}
Pennington, J., R.~Socher, and C.~Manning (2014, October).
\newblock {G}lo{V}e: Global vectors for word representation.
\newblock In {\em Proceedings of the 2014 Conference on Empirical Methods in
  Natural Language Processing ({EMNLP})}, Doha, Qatar, pp.\  1532--1543.
  Association for Computational Linguistics.

\bibitem[\protect\citeauthoryear{Peters, Ammar, Bhagavatula, and Power}{Peters
  et~al.}{2017}]{Peters2017SemisupervisedST}
Peters, M.~E., W.~Ammar, C.~Bhagavatula, and R.~Power (2017).
\newblock Semi-supervised sequence tagging with bidirectional language models.
\newblock In {\em ACL}.

\bibitem[\protect\citeauthoryear{Peters, Neumann, Iyyer, Gardner, Clark, Lee,
  and Zettlemoyer}{Peters et~al.}{2018}]{peters2018elmo}
Peters, M.~E., M.~Neumann, M.~Iyyer, M.~Gardner, C.~Clark, K.~Lee, and
  L.~Zettlemoyer (2018, June).
\newblock Deep contextualized word representations.
\newblock In {\em Proceedings of the 2018 Conference of the North {A}merican
  Chapter of the Association for Computational Linguistics: Human Language
  Technologies, Volume 1 (Long Papers)}, New Orleans, Louisiana, pp.\
  2227--2237. Association for Computational Linguistics.

\bibitem[\protect\citeauthoryear{Radford, Narasimhan, Salimans, and
  Sutskever}{Radford et~al.}{2018}]{radford2018improving}
Radford, A., K.~Narasimhan, T.~Salimans, and I.~Sutskever (2018).
\newblock Improving language understanding by generative pre-training.

\bibitem[\protect\citeauthoryear{Radford, Wu, Child, Luan, Amodei, and
  Sutskever}{Radford et~al.}{2019}]{radford2019language}
Radford, A., J.~Wu, R.~Child, D.~Luan, D.~Amodei, and I.~Sutskever (2019).
\newblock Language models are unsupervised multitask learners.
\newblock {\em OpenAI Blog\/}~{\em 1\/}(8), 9.

\bibitem[\protect\citeauthoryear{Rajpurkar, Jia, and Liang}{Rajpurkar
  et~al.}{2018}]{rajpurkar2016squadv2}
Rajpurkar, P., R.~Jia, and P.~Liang (2018, July).
\newblock Know what you don{'}t know: Unanswerable questions for {SQ}u{AD}.
\newblock In {\em Proceedings of the 56th Annual Meeting of the Association for
  Computational Linguistics (Volume 2: Short Papers)}, Melbourne, Australia,
  pp.\  784--789. Association for Computational Linguistics.

\bibitem[\protect\citeauthoryear{Rajpurkar, Zhang, Lopyrev, and
  Liang}{Rajpurkar et~al.}{2016}]{rajpurkar2016squadv1}
Rajpurkar, P., J.~Zhang, K.~Lopyrev, and P.~Liang (2016, November).
\newblock {SQ}u{AD}: 100,000+ questions for machine comprehension of text.
\newblock In {\em Proceedings of the 2016 Conference on Empirical Methods in
  Natural Language Processing}, Austin, Texas, pp.\  2383--2392. Association
  for Computational Linguistics.

\bibitem[\protect\citeauthoryear{Reimers and Gurevych}{Reimers and
  Gurevych}{2019}]{reimers2019sentencebert}
Reimers, N. and I.~Gurevych (2019, 11).
\newblock Sentence-bert: Sentence embeddings using siamese bert-networks.
\newblock In {\em Proceedings of the 2019 Conference on Empirical Methods in
  Natural Language Processing}. Association for Computational Linguistics.

\bibitem[\protect\citeauthoryear{{Salinas}, {Flunkert}, and
  {Gasthaus}}{{Salinas} et~al.}{2017}]{DeepAR2017}
{Salinas}, D., V.~{Flunkert}, and J.~{Gasthaus} (2017, April).
\newblock {DeepAR: Probabilistic Forecasting with Autoregressive Recurrent
  Networks}.
\newblock {\em arXiv e-prints\/}, arXiv:1704.04110.

\bibitem[\protect\citeauthoryear{Scarselli, Gori, Tsoi, Hagenbuchner, and
  Monfardini}{Scarselli et~al.}{2009}]{Scarselli2009GNN}
Scarselli, F., M.~Gori, A.~C. Tsoi, M.~Hagenbuchner, and G.~Monfardini (2009).
\newblock The graph neural network model.
\newblock {\em IEEE Transactions on Neural Networks\/}~{\em 20\/}(1), 61--80.

\bibitem[\protect\citeauthoryear{Socher, Chen, Manning, and Ng}{Socher
  et~al.}{2013}]{socher2013reasoning}
Socher, R., D.~Chen, C.~D. Manning, and A.~Ng (2013).
\newblock Reasoning with neural tensor networks for knowledge base completion.
\newblock In {\em NIPS}, pp.\  926--934.

\bibitem[\protect\citeauthoryear{Sperduti and Starita}{Sperduti and
  Starita}{1997}]{Sperduti1997GNN}
Sperduti, A. and A.~Starita (1997).
\newblock Supervised neural networks for the classification of structures.
\newblock {\em IEEE Transactions on Neural Networks\/}~{\em 8\/}(3), 714--735.

\bibitem[\protect\citeauthoryear{Tenney, Das, and Pavlick}{Tenney
  et~al.}{2019}]{tenney2019bert}
Tenney, I., D.~Das, and E.~Pavlick (2019, July).
\newblock {BERT} rediscovers the classical {NLP} pipeline.
\newblock In {\em Proceedings of the 57th Annual Meeting of the Association for
  Computational Linguistics}, Florence, Italy, pp.\  4593--4601. Association
  for Computational Linguistics.

\bibitem[\protect\citeauthoryear{Trivedi, Dai, Wang, and Song}{Trivedi
  et~al.}{2017}]{Trivedi2017KnowEvolve}
Trivedi, R., H.~Dai, Y.~Wang, and L.~Song (2017).
\newblock Know-evolve: Deep temporal reasoning for dynamic knowledge graphs.
\newblock In {\em Proceedings of the 34th International Conference on Machine
  Learning - Volume 70}, ICML'17, pp.\  3462–3471. JMLR.org.

\bibitem[\protect\citeauthoryear{Trouillon, Welbl, Riedel, Gaussier, and
  Bouchard}{Trouillon et~al.}{2016}]{Trouillon2016ComplEX}
Trouillon, T., J.~Welbl, S.~Riedel, E.~Gaussier, and G.~Bouchard (2016, 20--22
  Jun).
\newblock Complex embeddings for simple link prediction.
\newblock In M.~F. Balcan and K.~Q. Weinberger (Eds.), {\em Proceedings of The
  33rd International Conference on Machine Learning}, Volume~48 of {\em
  Proceedings of Machine Learning Research}, New York, New York, USA, pp.\
  2071--2080. PMLR.

\bibitem[\protect\citeauthoryear{Vaswani, Shazeer, Parmar, Uszkoreit, Jones,
  Gomez, Kaiser, and Polosukhin}{Vaswani et~al.}{2017}]{vaswani2017attention}
Vaswani, A., N.~Shazeer, N.~Parmar, J.~Uszkoreit, L.~Jones, A.~N. Gomez, L.~u.
  Kaiser, and I.~Polosukhin (2017).
\newblock Attention is all you need.
\newblock In I.~Guyon, U.~V. Luxburg, S.~Bengio, H.~Wallach, R.~Fergus,
  S.~Vishwanathan, and R.~Garnett (Eds.), {\em Advances in Neural Information
  Processing Systems}, Volume~30, pp.\  5998--6008. Curran Associates, Inc.

\bibitem[\protect\citeauthoryear{Veli\v{c}kovi\'{c}, Cucurull, Casanova,
  Romero, Liò, and Bengio}{Veli\v{c}kovi\'{c}
  et~al.}{2018}]{Velickovic2018model}
Veli\v{c}kovi\'{c}, P., G.~Cucurull, A.~Casanova, A.~Romero, P.~Liò, and
  Y.~Bengio (2018).
\newblock Graph attention networks.
\newblock ICLR 2018.

\bibitem[\protect\citeauthoryear{Wang, Singh, Michael, Hill, Levy, and
  Bowman}{Wang et~al.}{2019}]{wang2018glue}
Wang, A., A.~Singh, J.~Michael, F.~Hill, O.~Levy, and S.~R. Bowman (2019).
\newblock {GLUE}: A multi-task benchmark and analysis platform for natural
  language understanding.
\newblock In {\em International Conference on Learning Representations}.

\bibitem[\protect\citeauthoryear{Wang, Mao, Wang, and Guo}{Wang
  et~al.}{2017}]{wang2017knowledge}
Wang, Q., Z.~Mao, B.~Wang, and L.~Guo (2017).
\newblock Knowledge graph embedding: A survey of approaches and applications.
\newblock {\em IEEE TKDE\/}~{\em 29\/}(12), 2724--2743.

\bibitem[\protect\citeauthoryear{Wang and Li}{Wang and Li}{2016}]{wang2016text}
Wang, Z. and J.-Z. Li (2016).
\newblock Text-enhanced representation learning for knowledge graph.
\newblock In {\em IJCAI}, pp.\  1293--1299.

\bibitem[\protect\citeauthoryear{Wang, Zhang, Feng, and Chen}{Wang
  et~al.}{2014a}]{wang2014knowledgeb}
Wang, Z., J.~Zhang, J.~Feng, and Z.~Chen (2014a).
\newblock Knowledge graph and text jointly embedding.
\newblock In {\em EMNLP}.

\bibitem[\protect\citeauthoryear{Wang, Zhang, Feng, and Chen}{Wang
  et~al.}{2014b}]{wang2014knowledge}
Wang, Z., J.~Zhang, J.~Feng, and Z.~Chen (2014b).
\newblock Knowledge graph embedding by translating on hyperplanes.
\newblock In {\em AAAI}.

\bibitem[\protect\citeauthoryear{Xie, Liu, Jia, Luan, and Sun}{Xie
  et~al.}{2016}]{xie2016representation}
Xie, R., Z.~Liu, J.~Jia, H.~Luan, and M.~Sun (2016).
\newblock Representation learning of knowledge graphs with entity descriptions.
\newblock In {\em AAAI}.

\bibitem[\protect\citeauthoryear{Xu, Qiu, Chen, and Huang}{Xu
  et~al.}{2017}]{xu2017knowledge}
Xu, J., X.~Qiu, K.~Chen, and X.~Huang (2017).
\newblock Knowledge graph representation with jointly structural and textual
  encoding.
\newblock In {\em IJCAI}, pp.\  1318--1324.

\bibitem[\protect\citeauthoryear{Yang, Yih, He, Gao, and Deng}{Yang
  et~al.}{2015}]{yang2015DistMult}
Yang, B., W.-t. Yih, X.~He, J.~Gao, and L.~Deng (2015).
\newblock Embedding entities and relations for learning and inference in
  knowledge bases.
\newblock In {\em ICLR}.

\bibitem[\protect\citeauthoryear{Yao, Mao, and Luo}{Yao
  et~al.}{2019}]{yao2019kgbert}
Yao, L., C.~Mao, and Y.~Luo (2019).
\newblock Kg-bert: Bert for knowledge graph completion.
\newblock {\em ArXiv\/}~{\em abs/1909.03193}.

\bibitem[\protect\citeauthoryear{Zhang and Chen}{Zhang and
  Chen}{2018}]{Zhang2018LinkPrediction}
Zhang, M. and Y.~Chen (2018).
\newblock Link prediction based on graph neural networks.
\newblock In S.~Bengio, H.~Wallach, H.~Larochelle, K.~Grauman, N.~Cesa-Bianchi,
  and R.~Garnett (Eds.), {\em Advances in Neural Information Processing
  Systems}, Volume~31. Curran Associates, Inc.

\bibitem[\protect\citeauthoryear{Zhang, Li, Xia, Wang, and Jin}{Zhang
  et~al.}{2020}]{Zhang2020RevisitingGN}
Zhang, M., P.~Li, Y.~Xia, K.~Wang, and L.~Jin (2020).
\newblock Revisiting graph neural networks for link prediction.
\newblock {\em ArXiv\/}~{\em abs/2010.16103}.

\bibitem[\protect\citeauthoryear{Zhou, Cui, Hu, Zhang, Yang, Liu, Wang, Li, and
  Sun}{Zhou et~al.}{2020}]{Zhou2020GNN}
Zhou, J., G.~Cui, S.~Hu, Z.~Zhang, C.~Yang, Z.~Liu, L.~Wang, C.~Li, and M.~Sun
  (2020).
\newblock Graph neural networks: A review of methods and applications.
\newblock {\em AI Open\/}~{\em 1}, 57--81.

\bibitem[\protect\citeauthoryear{Şenay Ağca, Birge, Wang, and Wu}{Şenay
  Ağca et~al.}{2020}]{Aca2020PandemicSPLC}
Şenay Ağca, J.~Birge, Z.~Wang, and J.~Wu (2020).
\newblock The impact of covid-19 on supply chain credit risk.
\newblock {\em Risk Management eJournal\/}.

\end{thebibliography}

\newpage
\appendix



\section{Related Work}

A literature survey of knowledge graph embedding methods has been conducted by \citet{wang2017knowledge}. The methods can be categorized into two types of models. Say, we are given two entities $e_1$ and $e_2$ connected by a relationship $r$. The methods used in knowledge graph embeddings use a function to embed the entities ($h_{e_1}$ and $h_{e_2}$) and use some representation of $r$. Translational distance methods aim to minimize the distance between $h_{e_1} + g(r)$ and $h_{e_2}$, where $g(r)$ is some vector representation of the relationship $r$ \citep{bordes2013TransE,wang2014knowledge}. Semantic matching models, on the other hand, employ similarity-based scoring functions \citep{nickel2011RESCAL,yang2015DistMult}. DistMult \citep{yang2015DistMult} used a bilinear function $h_{e_1}^T A_r h_{e_2}$ to score the relationship where $A_r$ is a matrix representation of the relationship $r$. Our model architecture (\myref{Section}{sec:model_arch}) uses a similar method to score supplier-customer relationships.

The above methods, however, only use structural information observed in triples while different kinds of external information has been used to improve performance. For textual descriptions, methods vary from using word embeddings \citep{socher2013reasoning,wang2014knowledgeb,wang2016text} to convolutions \citep{xie2016representation} to attention \citep{xu2017knowledge,an2018accurate}. Similar to our method of handling company description, \citet{yao2019kgbert} used BERT to handle encoding of textual information. 

Another type of additional information that has been added is the graph structure (akin to the GNN we use in our architecture). In addition to the triples observed in a knowledge graph, path context (other paths connecting the two entities) have been used \citep{Lin2015PTransE,Guu2015Traversing,Feng2016GAKE} as well as neighboring triplets \citep{Feng2016GAKE,Jiang2021MultiRelGraph}.
Temporal information and information about when the relationship previously existed has also been incorporated \citep{Trivedi2017KnowEvolve}, similar to our model using historical supply chain data.

One main difference between our link prediction task and those of knowledge graph embeddings is we have only one type of link we are trying to predict.

In \myref{Appendix}{sec:gnn_background}, we discuss the relevant literature behind graph neural networks. Many of these methods can and have been used in link prediction. Some baselines include matrix factorization and node2vec \citep{grover2016node2vec}, though both are transductive methods. Another method for link prediction is SEAL \citep{Zhang2018LinkPrediction,Li2020DistanceEncoding} where a local enclosing subgraph is extracted around each target link and the nodes in each enclosing subgraph are labeled differently according to their distances to the source and target nodes \citep{Zhang2020RevisitingGN}.

\section{Graph Neural Networks}

\subsection{Background on Graph Neural Networks}\label{sec:gnn_background}

There has been many networks introduced to deal with arbitrarily structured graphs. Early work approached the problem using recurrent neural networks (RNNs). \citet{Frasconi1998GNN} and \citet{Sperduti1997GNN} used RNNs to process directed acyclic graphs; \citet{Gori2005GNN} and \citet{Scarselli2009GNN} developed a generalization of RNNs that can handle a large class of graphs. The method was further improved upon by \citet{Li2016GatedGNN} which used gated recurrent units (GRUs, \cite{Cho2014GRU}) in the propagation step.

In addition to the RNN approach, there has been research in generalizing convolutions to handle graphs. One method used convolutions in conjunction with a spectral representation of the graph such as the work in \citep{Bruna2013Spectral,Henaff2015DeepCN,Defferrard2016Convolutional,Kipf2017GCN}. Another method used convolutions directly on the graph \citep{Duvenaud2015Convolution,Atwood2016Diffusion,Monti2017Geometric,Hamilton2017model}; however, one challenge of these approaches is handling different sized neighborhoods and retaining the weight sharing property of convolutional networks. Though, most of these methods are transductive (learn a representation per node), GraphSAGE \citep{Hamilton2017model} used a inductive method to create node representations.

As attention-based mechanisms become the de facto method for many tasks with variable sized inputs \citep{bahdanau2014neural,vaswani2017attention,devlin2018bert}, \citet{Velickovic2018model} introduced Graph Attention Networks allowing for easier handling different size neighborhoods.

A thorough literature survey of graph neural networks has been conducted by \citet{Zhou2020GNN}.

\subsection{Graph Attentional Layer}\label{sec:gat_layer}

Among the methods that exist to process graphs, Graph Attention Networks \citep{Velickovic2018model} best fit our needs as we need a method that:

\begin{enumerate}
    \item Can be applied to graphs with differing structures since we would not want to \textit{have to} retrain our models anytime the data is updated; ideally, our model could continue to use new information as it arrives
    \item Is \textit{inductive} (as opposed to transductive), or in other words, can make predictions on nodes (companies) that the model has never seen before by using feature information to generate node embeddings
    \item Allows for flexibility in the face of different sized neighborhoods (variable number of neighbors)
\end{enumerate}

The first requirement precludes the usage of spectral convolutional methods \citep{Bruna2013Spectral,Henaff2015DeepCN,Defferrard2016Convolutional} since these methods require filters trained on specific spectral representations (Laplacian eigenbasis) which depends on the graph structure.
The second requirement does not necessarily preclude us from using many methods such as graph convolutional networks (GCNs, \cite{Kipf2017GCN}) as these often are extendable to the inductive case \citep{Hamilton2017GraphSAGE}. The third requirement leads us to use attention when aggregating neighborhoods instead of using other aggregators such as the mean over the feature vectors of the neighbors or using an RNN over the feature vectors. An important consideration when choosing an aggregator is to choose a function that is permutation-invariant since there is no inherent ordering of neighbors.

At a high level, the layer takes representations $\vec{h}_i$ for each node $i$ and a graph (the neighbors of each node) as input and outputs a new representation $\vec{h'}_i$ for each node. The new representation is formed by using an attention-based function whose inputs are the previous representation of the node and the previous representations of the neighbors of the node. 

More specifically, for some node $i$, we denote $N(i)$ as the neighbors of $i$. Using attention means that, for each node $i$, we compute a weight $w_{ij}$ by using its query $\vec{q}_i$ and the keys $\vec{k}_j$ for each of its neighbors $j$ and then output a weighted sum of the values $\sum_j w_{ij} \vec{v}_j$. For the graph attentional layer, this is achieved by:

\begin{align}
\begin{split}\label{eqn:gat_layer}
 \vec{q}_i = W_q h_i \qquad \vec{k}_j &= W_k h_j \qquad \vec{v}_k = W_v h_k
\\ a_{ij} &= \vec{q}_i \cdot \vec{k}_j
\\ w_{ij} &= \frac{a_{ij}}{\sum_{k \in N(i)} a_{ik}}
\\ \vec{h'}_{i} &= \sum_{j \in N(i)} w_{ij} \vec{v}_j
\end{split}
\end{align}

The choice in definition for $N(i)$ is arguably an arbitary choice made by the modeler, but often the choice comes down to whether or not to include a self-connection (is $i$ in $N(i)$)? In \citet{Velickovic2018model}, the self-connection is included; in our model, we do not include the self-connection (more details in \myref{Section}{sec:model_arch}).

\section{Comparing BERTs}\label{sec:bert}

\subsection{Background on Pretrained Language Models}

Learning representations of words and documents that are applicable to many tasks has been an active area of research for decades, including non-neural \citep{brown1992ngram,ando2005words,blitzer2006domainadap} and neural methods \citep{word2vec,pennington2014glove}. In early methods such as word2vec \citep{word2vec} and GloVe \citep{pennington2014glove}, the representations of the words lacked contextual information, e.g. though Apple is both a company and fruit, these methods would give the same embedding for apple, no matter the context.
These approaches were generalized to sentence embeddings \citep{Kiros2015Skip,logeswaran2018an} and paragraph embeddings \citep{quoc2014sentence2vec}.

ELMo \citep{peters2018elmo} and its predecessor \citep{Peters2017SemisupervisedST} generalized word embeddings by extracting context-sensitive information from a left-to-right and a right-to-left language model and concatenating the representations from the two models. 

Whereas ELMo used LSTMs, later work utilized Transformers \citep{vaswani2017attention} much more. GPT and its variants \citep{radford2018improving,radford2019language,brown2020language} use left-to-right language models. The method often used today to improve performance on NLP tasks has been BERT \citep{devlin2018bert} and its variants \citep{Liu2019RoBERTa,Lan2019ALBERT,reimers2019sentencebert}. BERT was trained using a loss to reconstruct masked tokens and a loss based on predicting whether, give a pair of sentences, does the second sentence follow the first in the original document. 

\subsection{BERT Experiments}

\begin{figure}[!tb]
\begin{center}

\begin{subfigure}{0.95\linewidth}
\centering
\centerline{\includegraphics[width=\columnwidth]{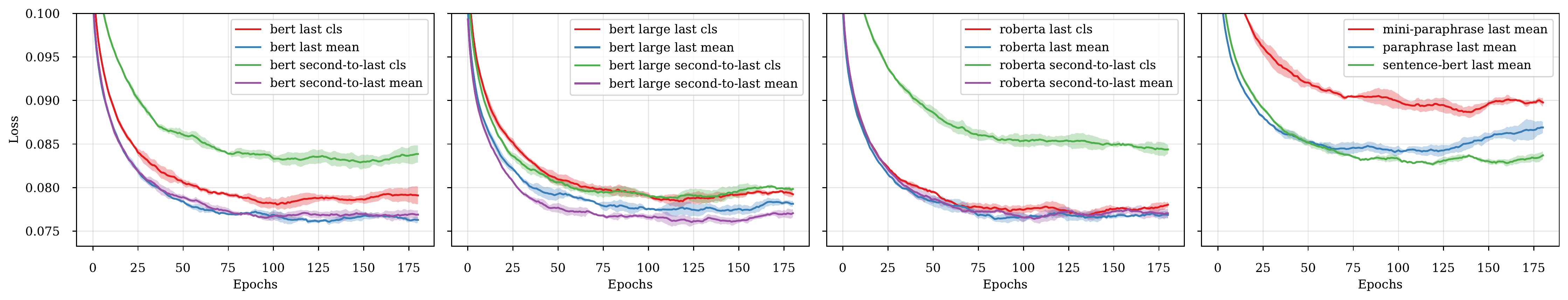}}
\caption{Comparison of BERTs without LSTMs}
\label{fig:compare_berts_frozen}
\end{subfigure}

\begin{subfigure}{0.95\linewidth}
\centering
\centerline{\includegraphics[width=\columnwidth]{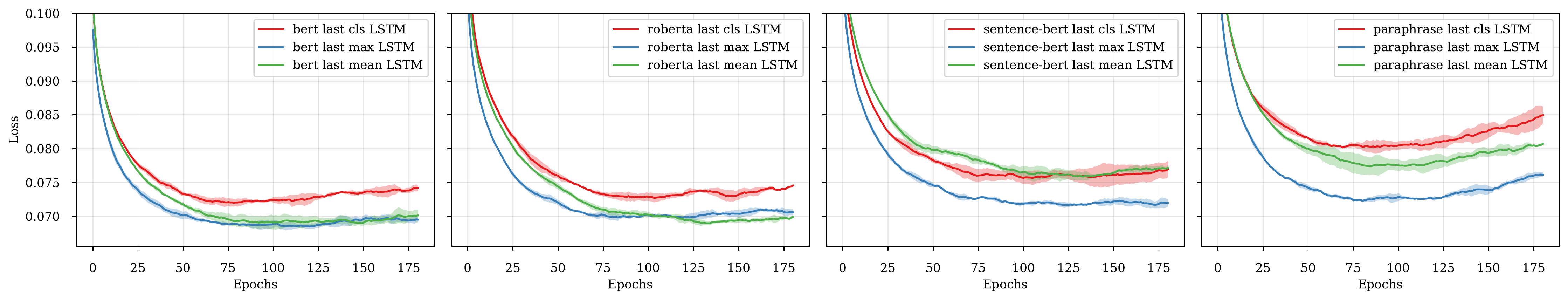}}
\caption{Comparison of BERTs with LSTMs}
\label{fig:compare_berts_bilstm}
\end{subfigure}

\begin{subfigure}{0.75\linewidth}
\centering
\centerline{\includegraphics[width=\columnwidth]{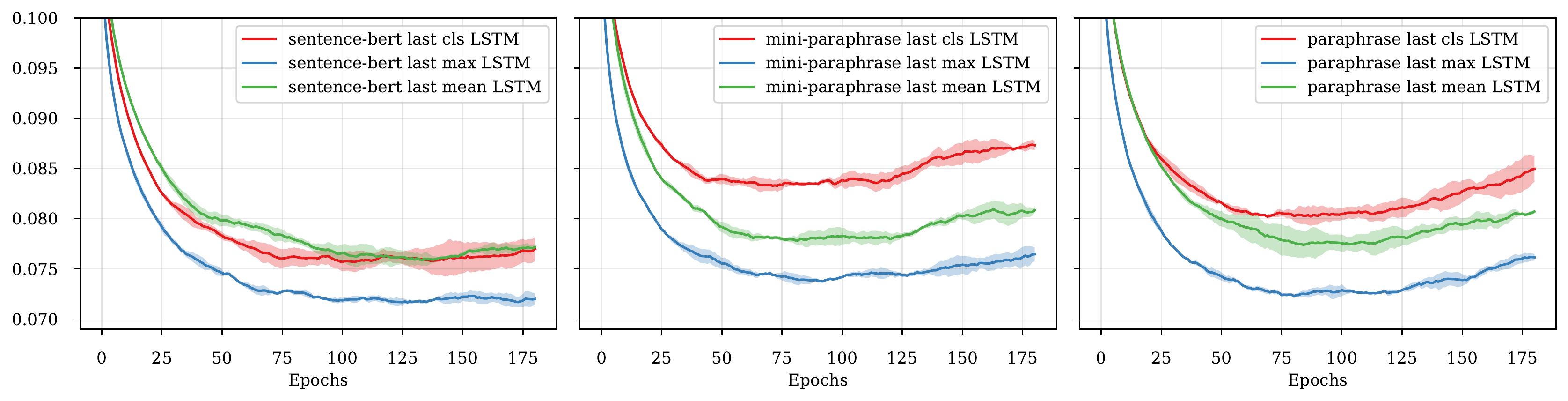}}
\caption{Comparison of Sentence BERTs with LSTMs}
\label{fig:sentence_berts_bilstm}
\end{subfigure}

\begin{subfigure}{0.5\linewidth}
\centering
\centerline{\includegraphics[width=\columnwidth]{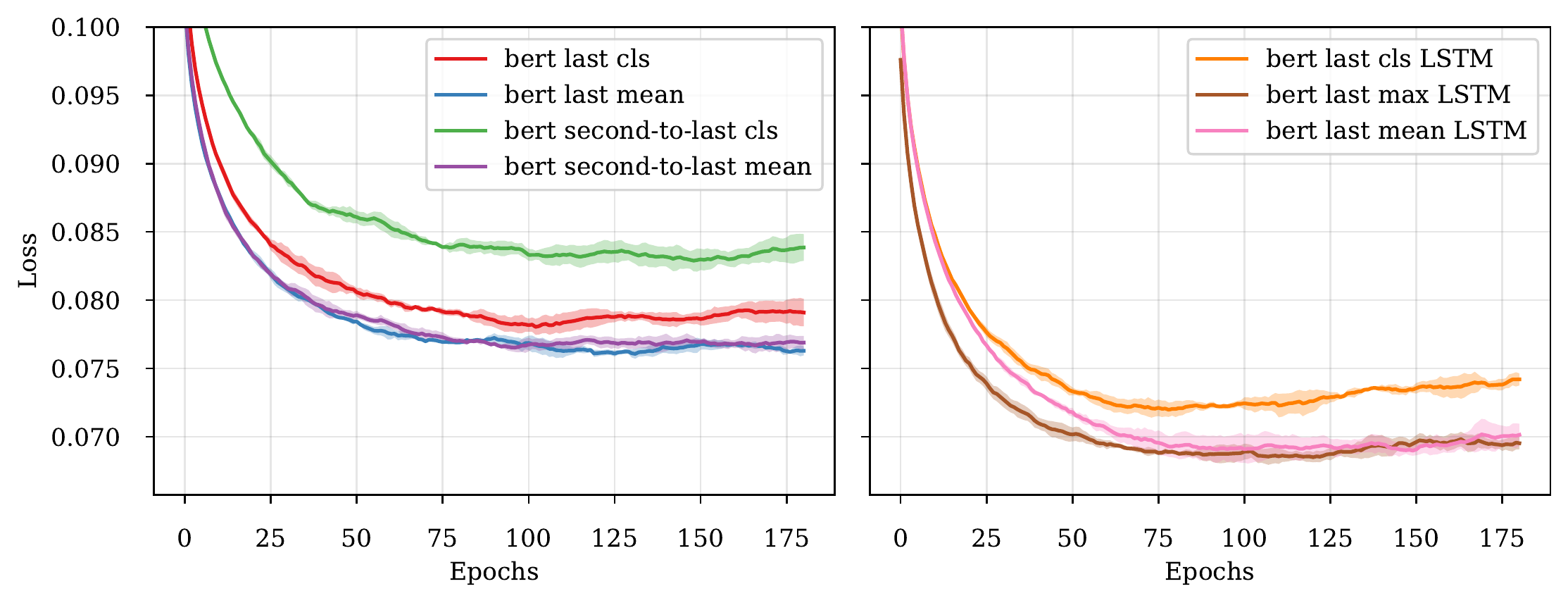}}
\caption{Comparison of BERT with and without LSTMs}
\label{fig:compare_berts_frozen_v_bilstm}
\end{subfigure}
\end{center}

\caption{Comparison of different BERT models and methods. For all the loss curves, we show a 20-epoch moving average and the standard deviation across three seeds. We do not show the performance without BERT in our plots as the loss is significantly larger ($0.1545 \pm 0.0002$ after 200 epochs).}
\label{fig:bert_comparisons}
\end{figure}

Though BERT has been shown to achieve strong performance in many NLP tasks \citep{devlin2018bert,Choi2020EvalBERT}, there has been variations in performance in downstream tasks depending on the BERT model used. For example, RoBERTa was shown to outperform BERT on GLUE (a collection of nine datasets for evaluating natural language understanding, \cite{wang2018glue}) and SQuAD (a question-answering dataset, \cite{rajpurkar2016squadv1,rajpurkar2016squadv2}); however, on sentence-level tasks such as Textual Similarity tasks, Sentence-BERTs outperform BERT \citep{reimers2019sentencebert}. Thus, in this section, we show the performance for supply chain link prediction across many different BERTs.

To simplify the task, we compare the performance between a model with only industry and country information versus a model with both as well as company description (processed using BERT).
Due to the computational cost of fine-tuning BERT in concurrence with supply chain dataset, we focused on the case of static embeddings (no fine-tuning of BERT). However, there are many methods to create static embeddings from BERT; we experimented with:
\begin{enumerate}

    \item Mean along sequence dimension of token embeddings (mean pooling)
    \item CLS vector

    In BERT models, sentences are padded with a start token, also know as CLS token. Using the CLS vector is the same as using the embedding for the first token. The justification for this is that when pretraining BERT, the CLS vector is used for the sentence level task (next sentence prediction). 

    \item Bidirectional-LSTM and aggregation

    Instead of fine-tuning, to add more expressivity, we used the token-level embeddings and passed those through a learnable bidirectional-LSTM. Then, we aggregate the processed token embeddings using one of the two aggregators above (mean pooling and CLS) or the maximum along sequence dimension of token embeddings (max pooling).

\end{enumerate}

We did not use the maximum along sequence dimension on the token embeddings from BERT as preliminary results showed this did not work well.

For the sentence BERTs, we only used the method that the model was intended to use; specifically, we focused on the sentence BERTs where the mean was used as the intended sentence embedding. This was only considered when not using a bidirectional LSTM; for the bidirectional LSTM, we tested all three aggregation methods.

Further, there have been studies on which layer to use when using BERT \citep{tenney2019bert,jawahar2019bert,liu2019bertlayers}; thus, we compare using the last layer versus the second to last layer.

The models we compared were: BERT-base, BERT-large, RoBERTa, and three variants of sentence BERT (denoted sentence BERT, paraphrase and mini-paraphase). Further, to ensure the difference in performance is coming from an understanding of company description, the suppliers in our validation set only included companies that were never seen in the training set. We ran each experiments three times.

In \myref{Figure}{fig:compare_berts_frozen}, we compare the performance on the validation set across the many different BERT models and aggregate methods, without LSTMs. We can see that the Sentence-BERTs significantly underperform the other BERTs. Further, though using the CLS token from the second to last layer in RoBERTa and BERT-base hurt performance, the performance for BERT-large is not significantly different from using the mean of the token embeddings. 
Further, for BERT-base and RoBERTa, the mean of the last layer performed the best whereas for BERT-large the mean of the second to last layer performed the best.

In \myref{Figure}{fig:compare_berts_bilstm}, we compare the performance using a bidirectional LSTM over the token embeddings. We can see that the best trainings of BERT and RoBERTA outperform sentence-BERT and its variants (\myref{Figure}{fig:sentence_berts_bilstm}). From \myref{Figure}{fig:compare_berts_frozen_v_bilstm}, we can see the LSTM gives a significant performance boost for BERT.
Further, though max pooling allows for faster learning in the BERT models, the performance gap between it and mean pooling is reduced with further training; in the sentence BERT models, max pooling gave the best performance when combined with bidirectional LSTMs.

In conclusion, due to the size of our supply chain model, we use the mean embeddings of the second to last layer of BERT-large instead of using a bidirectional LSTM with the token embeddings. Though, since our experiments showed an improvement with bidirectional LSTM, we leave it to future work to utilize this observation to its fully extent.

\section{Additional Evaluation}

\subsection{Stratified Performance}

\begin{figure}[!bt]
  \begin{minipage}[b]{0.48\textwidth}
    \centering
      \begin{tabular}{lcc}
\toprule
Supplier Industry &  Recall@100 &  Hit@20 \\
\midrule
Communications &    0.64 & 0.50 \\
Consumer Discretionary &    0.56 & 0.45 \\
Consumer Staples &    0.62 & 0.51 \\
Energy &    0.65 & 0.53 \\
Financials &    0.37 & 0.23 \\
Health Care &    0.56 & 0.44 \\
Industrials &    0.53 & 0.45 \\
Materials &    0.54 & 0.42 \\
Real Estate &    0.43 & 0.29 \\
Technology &    0.59 & 0.50 \\
Utilities &    0.54 & 0.42 \\
\bottomrule
\end{tabular}
\captionof{table}{Performance of our model on validation set across different industries on supplier side.}\label{tbl:eval_sup_industry}
    \end{minipage}
  \hfill
    \begin{minipage}[b]{0.48\textwidth}
    \centering
      \begin{tabular}{lcc}
\toprule
Customer Industry &  Recall@100 &  Hit@20 \\
\midrule
Communications &    0.68 & 0.49 \\
Consumer Discretionary &    0.55 & 0.37 \\
Consumer Staples &    0.61 & 0.40 \\
Energy &    0.72 & 0.50 \\
Financials &    0.39 & 0.20 \\
Health Care &    0.55 & 0.36 \\
Industrials &    0.49 & 0.31 \\
Materials &    0.49 & 0.28 \\
Real Estate &    0.36 & 0.17 \\
Technology &    0.59 & 0.39 \\
Utilities &    0.62 & 0.39 \\
\bottomrule
\end{tabular}
\captionof{table}{Performance of our model on validation set across different industries on customer side.}\label{tbl:eval_cus_industry}
    \end{minipage}
  \end{figure}

\begin{figure}[!bt]
\begin{minipage}[b]{0.48\textwidth}
    \centering
      \begin{tabular}{lcc}
      \toprule
      Supplier Country &  Recall@100 &  Hit@20 \\
      \midrule
      AU &    0.55 & 0.41 \\
      CA &    0.57 & 0.45 \\
      DE &    0.50 & 0.40 \\
      FR &    0.61 & 0.51 \\
      GB &    0.55 & 0.42 \\
      IN &    0.62 & 0.51 \\
      IT &    0.53 & 0.39 \\
      RU &    0.73 & 0.66 \\
      US &    0.66 & 0.53 \\
      \bottomrule
      \end{tabular}
\captionof{table}{Performance of our model on validation set across different countries on supplier side. We show performance on only nine countries.}\label{tbl:eval_sup_country}
    \end{minipage}
  \hfill
  \begin{minipage}[b]{0.48\textwidth}
    \centering
      \begin{tabular}{lcc}
      \toprule
      Customer Country &  Recall@100 &  Hit@20 \\
      \midrule
      AU &  0.53 & 0.33 \\
      CA &  0.51 & 0.30 \\
      DE &  0.50 & 0.28 \\
      FR &  0.70 & 0.51 \\
      GB &  0.63 & 0.41 \\
      IN &  0.63 & 0.46 \\
      IT &  0.53 & 0.35 \\
      RU &  0.74 & 0.59 \\
      US &  0.67 & 0.48 \\
      \bottomrule
      \end{tabular}
\captionof{table}{Performance of our model on validation set across different countries on customer side. We show performance on only nine countries.}\label{tbl:eval_cus_country}
    \end{minipage}
  \end{figure}

Though we show strong performance in aggregate (\myref{Section}{sec:evaluation}), we believe it to be a useful endeavor (especially when using a model in practice) to get a deeper understanding of performance across different subsets of the data. Specifically, we focus on the performance across different industries (\myref{Figure}{fig:bics_bucket}, \myref{Table}{tbl:eval_sup_industry}, \myref{Table}{tbl:eval_cus_industry}) as well as countries (\myref{Figure}{fig:country_bucket}, \myref{Table}{tbl:eval_sup_country}, \myref{Table}{tbl:eval_cus_country}). 

In \myref{Table}{tbl:eval_sup_country} and \myref{Table}{tbl:eval_cus_country}, we can see qualitatively that the performance is reasonably uniform; however, there is a bit of dip in performance for a few countries on the customer side. However, in \myref{Table}{tbl:eval_sup_industry} and \myref{Table}{tbl:eval_cus_industry}, we can see that there are industries the model struggles with more. Specifically, the largest hit in performance comes from Financials and Real Estate companies. This is to be expected as the amount of data in these industries is lower; further, many relationships in the real estate space can be from leasing to anyone, making it difficult to predict.

\begin{figure}[!tb]
\begin{center}

\begin{subfigure}{0.45\linewidth}
\centering
\centerline{\includegraphics[width=\columnwidth]{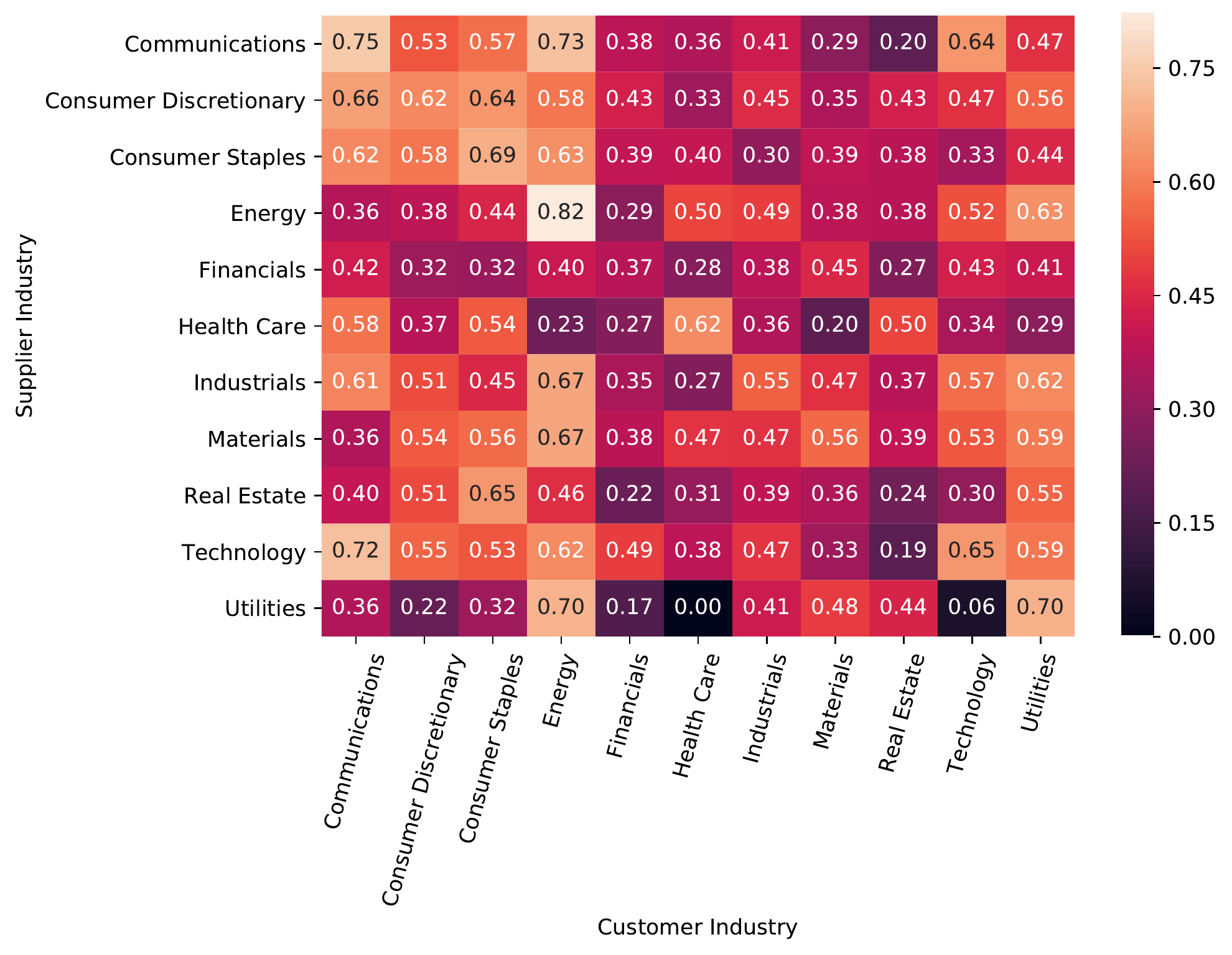}}
\caption{Recall@100}
\label{fig:recall_at_100_bics}
\end{subfigure}
\hfill
\begin{subfigure}{0.45\linewidth}
\centering
\centerline{\includegraphics[width=\columnwidth]{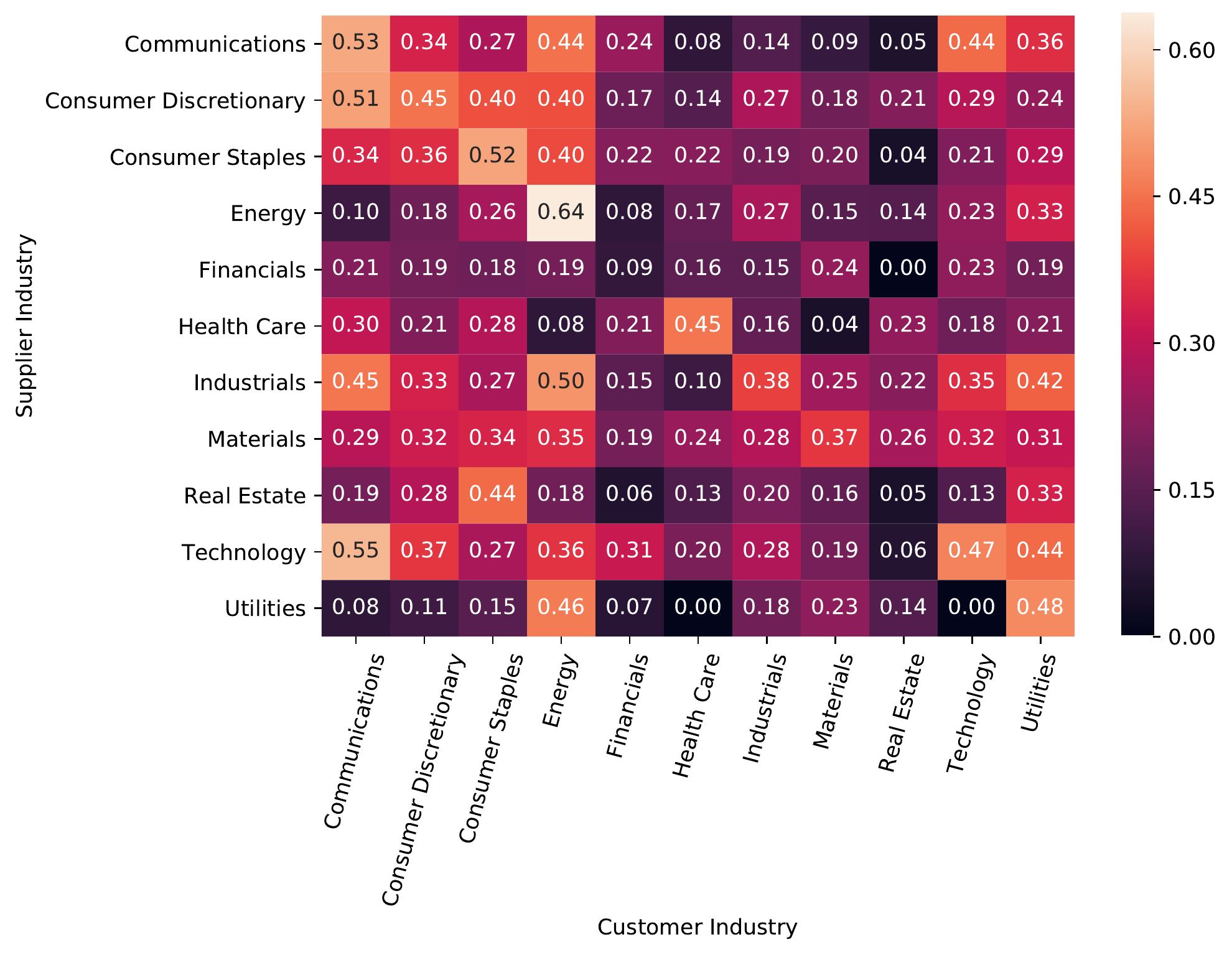}}
\caption{Hit@20}
\label{fig:hit_at_20_bics}
\end{subfigure}

\begin{subfigure}{0.45\linewidth}
\centering
\centerline{\includegraphics[width=\columnwidth]{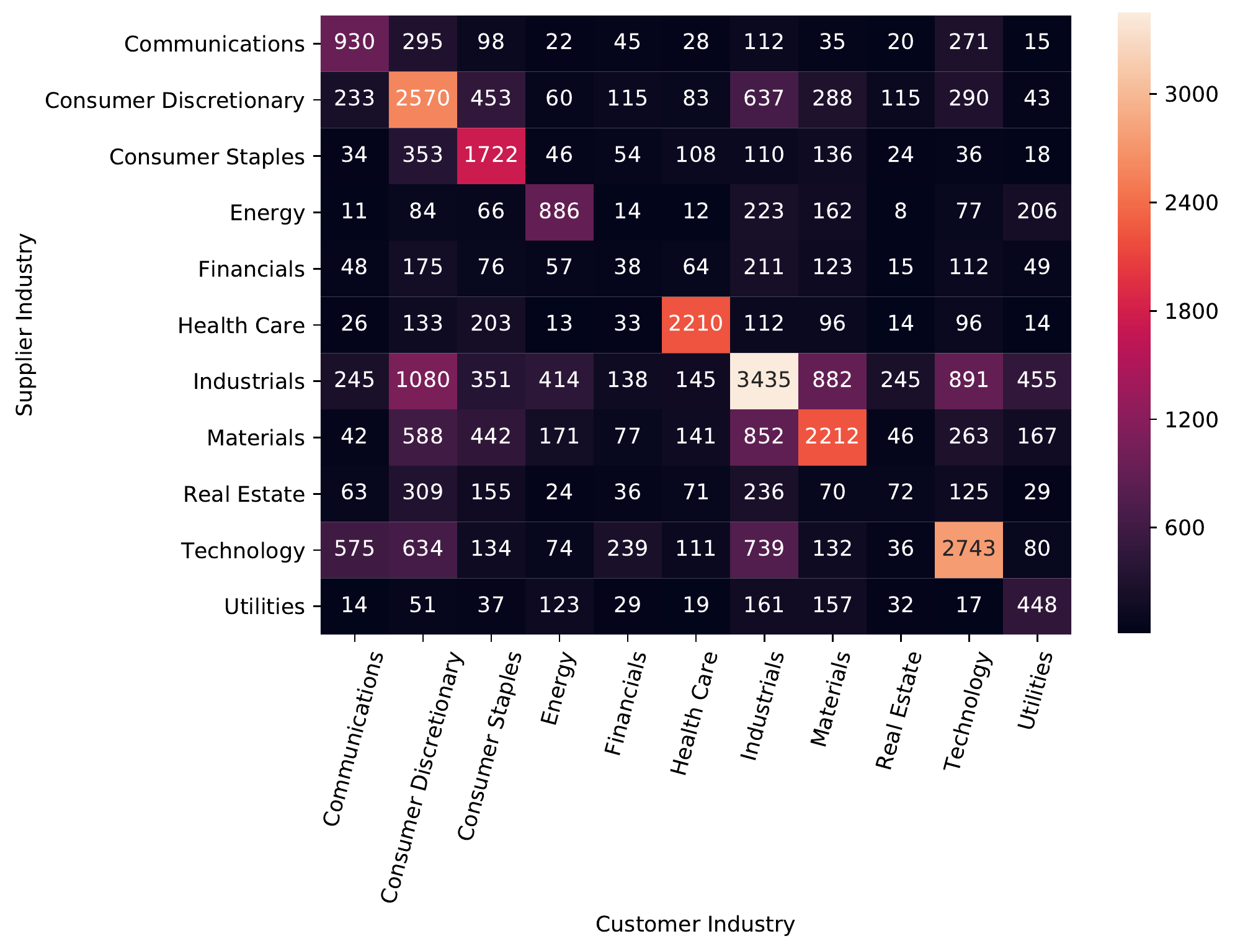}}
\caption{Counts}
\label{fig:counts_bics}
\end{subfigure}

\end{center}

\caption{Performance of our model on validation set across different industries. We further gives the counts in the corresponding to buckets to give a better sense if the performance metrics were computed with a sufficient number of observations. }
\label{fig:bics_bucket}
\end{figure}

\begin{figure}[!tb]
\begin{center}

\begin{subfigure}{0.45\linewidth}
\centering
\centerline{\includegraphics[width=\columnwidth]{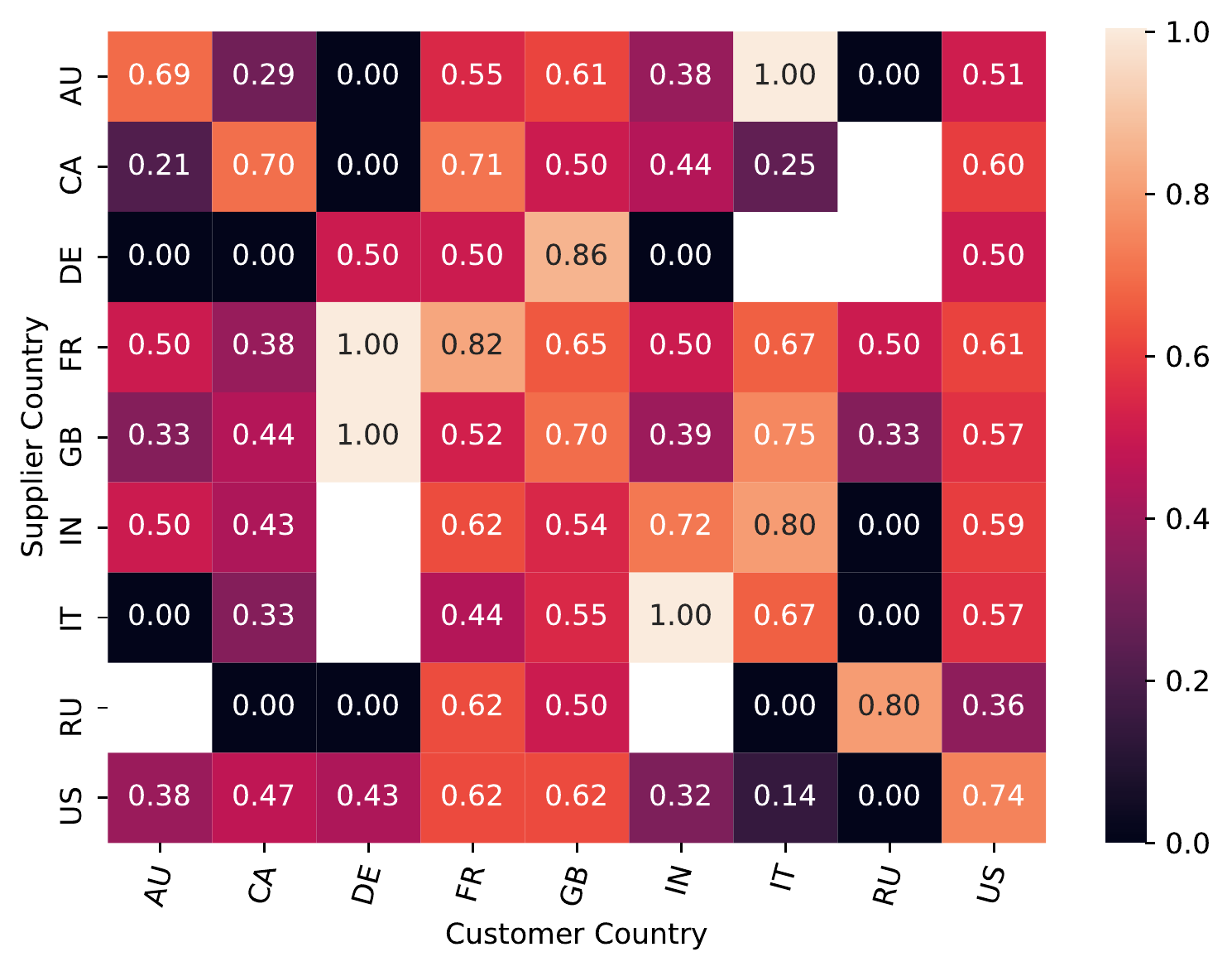}}
\caption{Recall@100}
\label{fig:recall_at_100_country}
\end{subfigure}
\hfill
\begin{subfigure}{0.45\linewidth}
\centering
\centerline{\includegraphics[width=\columnwidth]{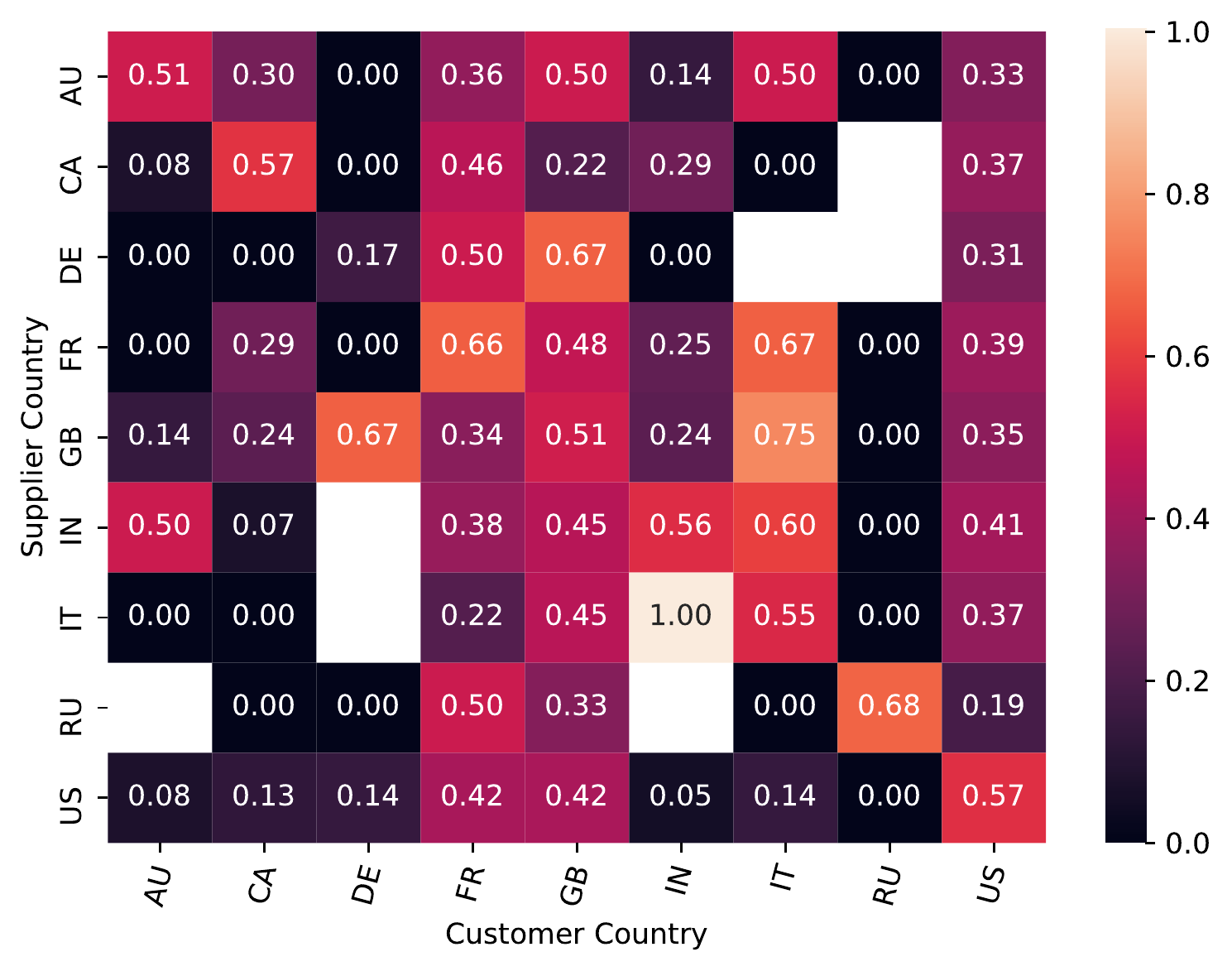}}
\caption{Hit@20}
\label{fig:hit_at_20_country}
\end{subfigure}

\begin{subfigure}{0.45\linewidth}
\centering
\centerline{\includegraphics[width=\columnwidth]{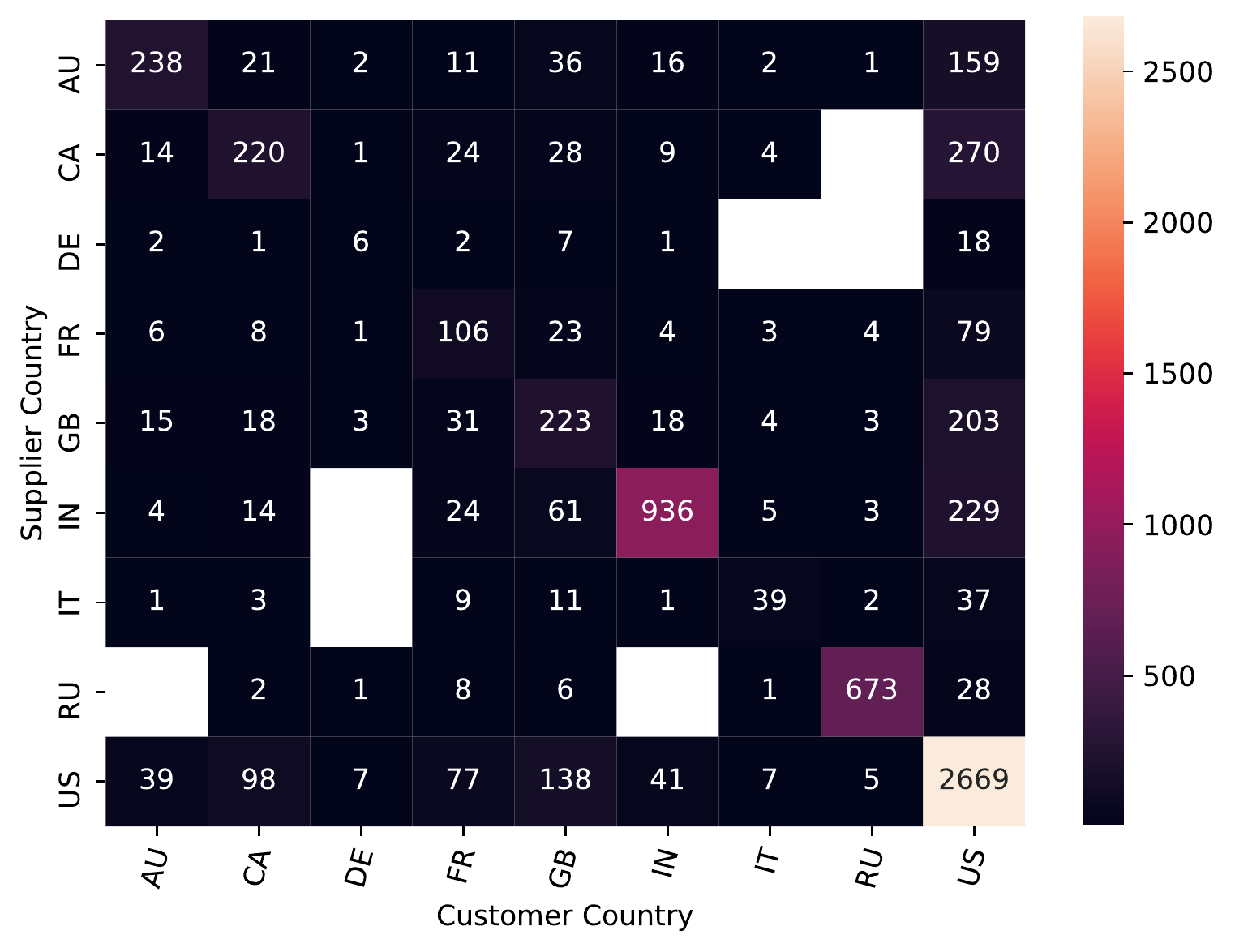}}
\caption{Counts}
\label{fig:counts_country}
\end{subfigure}

\end{center}

\caption{Performance of our model on validation set across different countries. We show performance on only nine countries. We further gives the counts in the corresponding to buckets to give a better sense if the performance metrics were computed with a sufficient number of observations. For example, though there is a 100\% recall from Italian to Indian companies, there is only one observation in that bucket; similarly, for Russian to Denmark companies, the recall is 0\% but also only has one observation.}
\label{fig:country_bucket}
\end{figure}

\end{document}